\documentclass[letterpaper, 10 pt, conference]{ieeeconf}  

\IEEEoverridecommandlockouts                              

\usepackage{tikz}
\usepackage{caption}
\usepackage{booktabs}
\usepackage{pgfplots}
\usepackage{adjustbox}
\usepackage{fancyhdr}

\usepackage{comment}
\usepackage{siunitx}
\usepackage{relsize}
\usepackage{ifthen}
\usepackage[colorinlistoftodos]{todonotes}

\usepackage[caption=false]{subfig}

\usepackage{cite}
\usepackage[vlined,ruled,linesnumbered]{algorithm2e}
\usepackage{graphics} 
\usepackage{rotating}
\usepackage{color}
\usepackage{enumerate}
\usepackage[T1]{fontenc}
\usepackage{psfrag}
\usepackage{epsfig} 
\usepackage{hyperref}
\usepackage{booktabs}
\usepackage{graphicx,url}
\usepackage{multirow}
\usepackage{array}
\usepackage{latexsym}
\usepackage{amsfonts}
\usepackage{amsmath}
\usepackage{amssymb}
\usepackage{xstring}
\usepackage[noend]{algorithmic}
\usepackage{multirow}
\usepackage{xcolor}
\usepackage{prettyref}
\usepackage{flexisym}
\usepackage{bigdelim}
\usepackage{breqn} 
\usepackage{listings}
\let\labelindent\relax
\usepackage{enumitem}
\usepackage{xspace}
\usepackage{bm}
\graphicspath{{./figures/}}
\usepackage{tikz}
\usetikzlibrary{matrix,calc}

\usepackage{mdwlist}

\let\itemize\undefined
\makecompactlist{itemize}{stditemize}

\newrefformat{prob}{Problem\,\ref{#1}}
\newrefformat{def}{Definition\,\ref{#1}}
\newrefformat{sec}{Section\,\ref{#1}}
\newrefformat{sub}{Section\,\ref{#1}}
\newrefformat{prop}{Proposition\,\ref{#1}}
\newrefformat{app}{Appendix\,\ref{#1}}
\newrefformat{alg}{Algorithm\,\ref{#1}}
\newrefformat{cor}{Corollary\,\ref{#1}}
\newrefformat{thm}{Theorem\,\ref{#1}}
\newrefformat{lem}{Lemma\,\ref{#1}}
\newrefformat{fig}{Fig.\,\ref{#1}}
\newrefformat{tab}{Table\,\ref{#1}}

\newcommand{\bdmath}{\begin{dmath}}
\newcommand{\edmath}{\end{dmath}}
\newcommand{\beq}{\begin{equation}}
\newcommand{\eeq}{\end{equation}}
\newcommand{\bdm}{\begin{displaymath}}
\newcommand{\edm}{\end{displaymath}}
\newcommand{\bea}{\begin{eqnarray}}
\newcommand{\eea}{\end{eqnarray}}
\newcommand{\beal}{\beq \begin{array}{ll}}
\newcommand{\eeal}{\end{array} \eeq}
\newcommand{\beas}{\begin{eqnarray*}}
\newcommand{\eeas}{\end{eqnarray*}}
\newcommand{\ba}{\begin{array}}
\newcommand{\ea}{\end{array}}
\newcommand{\bit}{\begin{itemize}}
\newcommand{\eit}{\end{itemize}}
\newcommand{\ben}{\begin{enumerate}}
\newcommand{\een}{\end{enumerate}}

\newcommand{\calE}{{\cal E}}

\newcommand{\calG}{{\cal G}}

\newcommand{\calV}{{\cal V}}

\newcommand{\etal}{\emph{et~al.}\xspace}

\newcommand{\eg}{\emph{e.g.,}\xspace}
\newcommand{\ie}{\emph{i.e.,}\xspace}
\newcommand{\myParagraph}[1]{{\bf #1.}\xspace}

\newcommand{\hide}[1]{}

\newcommand{\highlight}[1]{#1} 

\newcommand{\hiddenText}{{\color{gray} hidden text.}}
\newcommand{\hideWithText}[1]{\hiddenText}

\newcommand{\blue}[1]{{\color{blue}#1}}

\newcommand{\linkToPdf}[1]{\href{#1}{\blue{(pdf)}}}
\newcommand{\linkToPpt}[1]{\href{#1}{\blue{(ppt)}}}
\newcommand{\linkToCode}[1]{\href{#1}{\blue{(code)}}}
\newcommand{\linkToWeb}[1]{\href{#1}{\blue{(web)}}}
\newcommand{\linkToVideo}[1]{\href{#1}{\blue{(video)}}}
\newcommand{\linkToMedia}[1]{\href{#1}{\blue{(media)}}}
\newcommand{\award}[1]{\xspace} 

\usepackage{tikz}
\usetikzlibrary{positioning}

\def\cma{0.35}
\def\cmb{0.3}

\newlength{\layerwidth}
\newlength{\layerheight}
\definecolor{cscene}{rgb}{1,1,1}
\definecolor{c2}{rgb}{0.15,0.278,0.33}
\definecolor{c3}{rgb}{0.16,0.62,0.56}
\definecolor{c4}{rgb}{0.91,0.77,0.42}
\definecolor{cagent}{rgb}{0.91,0.44,0.32}

\definecolor{intercolor}{rgb}{0.5,0.5,0.5}

\newlength{\scene}
\newlength{\layertwo}
\newlength{\layerthree}
\newlength{\layerfour}
\newlength{\layeragent}
\pgfmathsetmacro{\minimalwidth}{0.2}
\pgfmathsetmacro{\nodeinnersep}{0.0}

\tikzstyle node2=[circle, draw, fill=c2, fill opacity=0.5, cm={1,0,\cma,\cmb,(0,0)}, minimum width=\minimalwidth*1cm,inner sep=0]
\tikzstyle node3=[circle, draw, fill=c3, fill opacity=0.5, cm={1,0,\cma,\cmb,(0,0)}, minimum width=\minimalwidth*1cm,inner sep=0]
\tikzstyle node4=[circle, draw, fill=c4, fill opacity=0.5, cm={1,0,\cma,\cmb,(0,0)}, minimum width=\minimalwidth*1cm,inner sep=0]
\tikzstyle nodeagent=[circle, draw, fill=cagent, fill opacity=0.75, cm={1,0,\cma,\cmb,(0,0)}, minimum width=\minimalwidth*1cm,inner sep=0]

\tikzstyle inter=[dashed, color=intercolor, opacity=0.7]
\tikzstyle same=[solid, opacity=0.7]
\tikzstyle highlight=[line width=2pt, dashed, color=cagent, opacity=0.7]

\newcommand{\DSG}{DSG\xspace}

\newcommand{\DSGt}{$\calG_{{\tiny\text{s}}}^t$\xspace}

\newcommand{\GAL}[1]{\calG_{{\tiny\text{a}}}^{#1}}
\newcommand{\nodesAL}[1]{\calV_{{\tiny\text{a}}}^{#1}}
\newcommand{\edgesAL}[1]{\calE_{{\tiny\text{a}}}^{#1}}
\newcommand{\graphObservationAtt}{\calG^t}
\newcommand{\nodeObservationAtt}{\calV^t}
\newcommand{\edgeObservationAtt}{\calE^t}

\newcommand{\optional}[1]{}
\newcommand{\veryOptional}[1]{}
\newcommand{\hideout}[1]{}

\usepackage{xcolor}
\usepackage{soul} 

\newcommand{\citet}[1]{\highlight{\cite{#1}}}
\newcommand{\citep}[1]{\cite{#1}}

\fancypagestyle{firstpage}
{
  \vspace{-3mm}
  \fancyhead[C]{\vspace{-10mm}Accepted for publication at ICRA 2022. Please cite as: \\
    Z. Ravichandran, L. Peng, N. Hughes, J. D. Griffith, and L. Carlone, 
    ``Hierarchical Representations and Explicit Memory: Learning Effective Navigation Policies on 3D Scene Graph using Graph Neural Networks," \textit{ICRA}, 2022.}
}

\title{\LARGE \bf
\vspace{3mm}Hierarchical Representations and Explicit Memory: 
Learning Effective Navigation Policies on 3D Scene Graphs using Graph Neural Networks
\vspace{-3mm}
}

\author{Zachary Ravichandran$^{1}$, Lisa Peng$^{2}$, Nathan Hughes$^{2}$, J. Daniel Griffith$^{1}$, and Luca Carlone$^{2}$
\thanks{
  DISTRIBUTION STATEMENT A. Approved for public release. Distribution is unlimited.
}%
\thanks{
  This material is based upon work supported by the Under Secretary of Defense for Research and Engineering under Air Force Contract No. FA8702-15-D-0001. Any opinions, findings, conclusions or recommendations expressed in this material are those of the author(s) and do not necessarily reflect the views of the Under Secretary of Defense for Research and Engineering.
}%
\thanks{$^{1}$Lincoln Laboratory, Massachusetts Institute of Technology, USA, \{zachary.ravichandran,dan.griffith\}@ll.mit.edu}%
\thanks{$^{2}$Laboratory for Information \& Decision Systems, Massachusetts Institute of Technology, USA, \{lisapeng,na26933,lcarlone\}@mit.edu}%
}

\begin{document}

\maketitle
\pagestyle{plain}
\thispagestyle{firstpage}



\begin{abstract}
    Representations are
    crucial for a robot to learn effective navigation policies.
    Recent work has shown that mid-level perceptual abstractions, such as depth estimates or 2D semantic segmentation, lead to more effective policies when provided as observations in place of raw sensor data (\eg~RGB images).
    However, such policies must still learn latent three-dimensional scene properties from mid-level abstractions.
    In contrast, high-level, hierarchical 
    representations such as \emph{3D scene graphs} explicitly provide a scene's geometry, topology, and semantics, making them compelling representations for navigation.
    In this work, we present a reinforcement learning framework that leverages high-level hierarchical representations to learn navigation policies.
    Towards this goal, we propose a graph neural network architecture and show how to embed a 3D scene graph into an agent-centric feature space,
    which enables the robot to learn policies \highlight{that map 3D scene graphs to a platform-agnostic control space (\eg go straight, turn left).}
    For each node in the scene graph, our method uses features that capture occupancy and semantic content, while explicitly
    retaining memory of the robot trajectory.
    We demonstrate the effectiveness of our method against commonly used visuomotor policies in a challenging multi-object search task.
    These experiments and supporting ablation studies show that our method leads to
    more effective object search behaviors,
    exhibits improved long-term memory,
    and 
    successfully leverages hierarchical information to guide its navigation objectives.
\end{abstract}


\section*{\highlight{Supplementary Material}}
\label{sec:supplemental}
\begin{center}
    \highlight{
        \small{
            Software: \href{https://github.com/MIT-TESSE/dsg-rl}{https://github.com/MIT-TESSE/dsg-rl} \\
            Video: \href{https://youtu.be/x4LM-g3-uaY}{https://youtu.be/x4LM-g3-uaY}
        }
    }
\end{center}


\section{Introduction}
\label{sec:intro}

A key objective when learning robot navigation policies is to design techniques that are sample efficient and generalize to unseen spaces.
A large body of reinforcement learning research focused on learning policies that directly map raw observations to actions~\cite{li2017deep,mnih2013playing}.
More recent research 
has shown that using mid-level representations, which provide an abstracted view of the world (\eg depth or 2D semantic segmentation) as inputs to a policy, is often more sample efficient and generalizes better~\citep{Sax2019corl-MidLevelReps,Chen2020corl-MidLevelReps,Zhou2019ScienceRobotics,Muller2018Corl,Mousavian2019ICRA}.
Similarly, early deep reinforcement learning research involved stacking previous observations to represent history and retain memory~\citep{mnih2013playing,wu2016training}.
However, more recent work achieves high-quality policies by relying on recurrent neural network architectures to implicitly learn an environment state, or using explicit spatially-structured memory similar to the ones produced by localization and mapping methods~\citep{shehroze2016doom,zhang2017neural,Chaplot2020CVPR}.

\begin{figure}[t!]
    \centering
    \vspace{3mm}
    \resizebox{1\columnwidth}{!}{\input{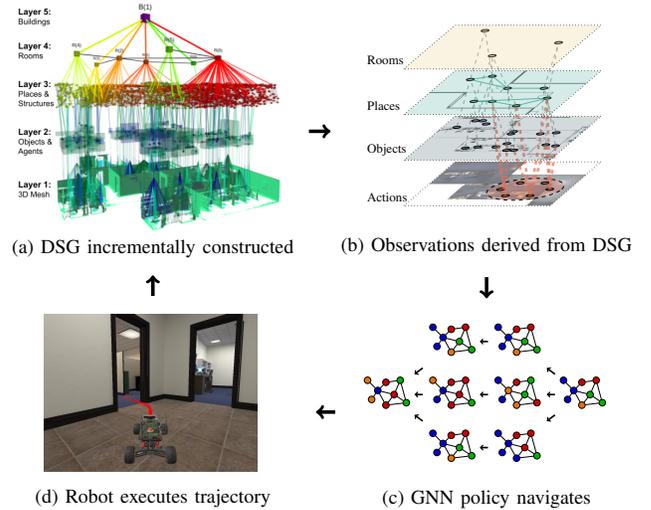}}
    \vspace{-4mm}
    \caption{\label{fig:graph-obs}
    \highlight{We design reinforcement-learning-based policies that map 3D Dynamic Scene Graphs (DSGs) directly to a platform-agnostic control space. 
    An incrementally constructed DSG (a) is augmented with an \textit{Action Layer} (b), 
    which encodes high-level DSG information into a robot-centric space. 
    Policies based on graph neural networks (GNNs) (c) then map these
    observations to actions (d).
    \vspace{-6mm}
    }}
\end{figure}

While the use of mid-level representations enables the design of more effective navigation policies,
these policies still require a robot to learn how to map mid-level representations to more appropriate higher-level concepts (\eg occupancy, topology of the environment, correlation between phenomena of interest and 3D locations) to support navigation.
These mappings can be very complex.
Similarly, the robot has to learn how to retain memory
in a way that is useful for the given task.
If a robot already had access to higher-level representations and appropriate history, then it seems likely that training could be even more sample efficient and more generalizable to unseen spaces.

Recent advances in robot perception and computer vision provide unprecedented opportunities to build explicit spatial memory
that yields a concise high-level description of the environment.
Modern SLAM systems can build 3D metric-semantic maps
in real-time from semantically labeled images~\cite{Tateno17cvpr-CNN-SLAM,Lianos18eccv-VSO,Dong17cvpr-XVIO,Behley19iccv-semanticKitti,McCormac17icra-semanticFusion,Zheng19arxiv-metricSemantic,Tateno15iros-metricSemantic,Li16iros-metricSemantic,McCormac183dv-fusion++,Runz18ismar-maskfusion,Runz17icra-cofusion,Xu19icra-midFusion,Rosinol20icra-Kimera,Grinvald19ral-voxbloxpp,Rosinol20rss-dynamicSceneGraphs}.
More recent algorithms for spatial perception can construct \emph{3D scene graphs} from intermediate representations~\citep{Armeni19iccv-3DsceneGraphs} or directly from sensor data~\citep{Rosinol20rss-dynamicSceneGraphs,Rosinol21arxiv-Kimera}.
3D scene graphs are powerful high-level representations of an environment, that capture its geometry, topology, and semantics.
The nodes in a 3D scene graph correspond to entities in the environment (\eg
objects, places, rooms, buildings), while edges describe relations
among these entities (\eg ``object $i$ is in room $j$'').
A scene graph encodes the environment at multiple
levels of abstraction, which is useful to capture task-dependent regularities (\eg a given object is more likely to
be present in certain rooms than others).

\myParagraph{Contribution} We present a reinforcement learning framework for learning navigation policies on 3D scene graphs.
In particular, we use the \emph{3D Dynamic Scene Graph} (DSG) \highlight{of Rosinol \etal} \citet{Rosinol20rss-dynamicSceneGraphs,Rosinol21arxiv-Kimera}, which includes
a topological map of places and is specifically designed for navigation.
Our approach, described in Section~\ref{sec:contribution}, uses graph neural networks to compute a navigation policy.
A key element in our proposal is a mechanism to embed a globally-defined DSG into an agent-centric feature space.
This mechanism enables us to learn \highlight{platform-agnostic policies for navigation (\eg go straight, turn) from DSG-derived observations.}
In contrast, most related works using graphs for robot navigation 
only predict subgoals over the graph (and then rely on an external low-level controller),
require a priori knowledge of the environment, or are tailored to a specific task~\citep{kurenkov2021icra-HMS,Sunderhauf2020arxiv-keys,ZhuCoRL2020,NguyenCoRL2020}.

We demonstrate our approach on a challenging multi-object search task in a simulated office environment (Section~\ref{sec:experiments}). The robot needs to search a previously unseen environment, while prioritizing search in areas where objects are more likely to be found.
Our results show that our approach outperforms popular visuomotor navigation policies.
Our experiments show that the hierarchical information in the \DSG helps the robot focus its search where a target object is more likely to be found. In particular, if we remove DSG nodes representing rooms {and objects}, then the number of target objects that the robot finds is reduced; this implies that the robot learns to correlate rooms and objects to where target objects are most likely found.
Our experiments also show that retaining explicit memory via the \DSG helps the robot avoid revisiting areas it has already searched.
We intend to release our software and simulator.


\section{Related Work}
\label{sec:relatedWork}

\myParagraph{Intermediate Representations in Reinforcement Learning} Sax \etal \citet{Sax2019corl-MidLevelReps} and Zhou \etal \citet{Zhou2019ScienceRobotics} show that using mid-level representations (\eg depth, semantic segmentation, surface normals) in lieu of raw sensor data as observations to learned policies improves sample efficiency, performance, and generalization of learned policies. 
Chen \etal \citet{Chen2020corl-MidLevelReps} experimentally validate these claims in sim-to-sim manipulation and sim-to-real navigation tasks. 
Muller \etal \citet{Muller2018Corl} use semantic segmentation to train a navigation policy that was tested on a physical platform. 
Mousavian \etal \citet{Mousavian2019ICRA} use semantic segmentation and object detection estimates as inputs to a target-driven navigation policy.
Our work extends these findings by showing that \emph{high-level} abstractions, such as 3D scene graphs, can further boost performance by explicitly encoding geometric, topological, and semantic information. 

\myParagraph{Semantic Knowledge for Robot Planning} Relationships provided through semantic knowledge embeddings have been shown to aid robot policy generalization to novel tasks and objects~\cite{Fulda2017corl, thomasonaaai18, darunaicra2019}. 
Semantic priors have improved robot policy performance on tasks including target-driven navigation~\cite{Yang2019icml},
manipulation~\cite{MuraliCoRL2020, Scaliseiros19}, and task execution from natural language instructions~\cite{Arkinijrr2020}. 
Several works associate object detections with prior embeddings to improve object search policies~\cite{DuECCV2020, Druon2020RAL}. 
Chaplot \etal \citet{chaplot2020object} propose a modular object search method that first builds an internal semantic map; a policy then uses that map to predict goal locations executed by a model-based planner.
Qiu \etal \citet{Qui2020CORL} fuse spatial information from an object detector and semantic information from a knowledge graph to create a hierarchical representation that captures object relations for target search.
Though our representations capture metric-semantic information like~\cite{chaplot2020object}, they are built by 3D scene graphs and explicitly model scene hierarchy and trajectory history.
While we do not use existing semantic corpuses (\eg Visual Genome~\cite{Krishnaijcv2017}), the representations we use provide semantic relationships over entities in the environment (\eg hallways, offices).

\myParagraph{Memory in Reinforcement Learning} Maintaining a robust notion of memory has been a limiting factor for learned policies in tasks such as exploration~\cite{ChenICLR2019} and navigation~\cite{MirowskiICLR2016}.
Novel policy architectures have been proposed as a means of preserving memory~\cite{shehroze2016doom, zhang2017neural,chaplot2020learning, ParisottoICLR2018}.
Gupta \etal \citet{Gupta2020IJCV} propose an architecture composed of 2D mapping and planning modules for navigation. 
Chaplot \etal \citet{Chaplot2020CVPR} propose learning semantic and geometric representations in a graphical structure. 
Wu \etal \citet{Wu2019ICCV} introduce Bayesian Relational Memory (BRM), a topological and semantic representation for navigation policies.
Building external representations of the environment has also been used as an alternative approach to capture memory~\cite{Chen2019RSS}. 
Beeching \etal \citet{BeechingECCV2020} learn an internal graph representation constructed during a pre-episode rollout to inform a low-level 
point-navigation policy. 
Savinov \etal \citet{SavinovICLR2018} construct a topological map of an environment during an explorative rollout phase.
%
\highlight{These latter two works~\cite{BeechingECCV2020, SavinovICLR2018}} are most similar to our method.
However, the \highlight{graphs} we use capture multiple levels of abstraction, are not task specific, and can be constructed online.

\myParagraph{Graph Neural Networks for Reinforcement Learning}
RL frameworks have used Graph Neural Networks (GNNs) to learn policies for tasks including robot exploration~\cite{Chen2020GCNNExploration}, multi-agent coordination~\cite{Jiang2020GraphConvRL}, and active learning~\cite{hu2020GraphPolicyNetwork}.
Gammelli \etal \citet{gammelli2021graph} use GNNs to learn vehicle routing policies over transportation networks.
Wang \etal \citet{Wang2018NerveNet} model multi-joint robots through graphical structures and
use these graph representations to learn continuous control policies.
Sunderhauf \citet{Sunderhauf2020arxiv-keys} uses semantic word embeddings to learn a policy that finds unmapped objects in graph representations of a scene. 
These works employ graphs that are ``flat'' (rather than hierarchical) and are generated offline. Moreover, they 
learn polices that operate over such graphs (\eg the robot plans over nodes in the graph rather than over low-level actions).
Our method instead uses dynamically constructed graphs that capture multiple levels of abstraction, and directly learns an agent-centric navigation policy.

\myParagraph{Scene Graphs} 2D scene graphs have been widely used in computer vision to describe object relationships in images~\cite{Krishnaijcv2017}. These representations have proven useful for tasks including image retrieval~\cite{Johnson_2015_CVPR, Schroeder_2020_CVPR_Workshops, Wang_2020_WACV}, image generation~\cite{johnson2018image, herzigeccv2020, Tripathi2019UsingSG}, and visual question answering~\cite{Ghosh2019GeneratingNL, Shi_2019_CVPR}. 
Much attention has been devoted to inferring scene graphs from images~\cite{xu2017scenegraph, yang2018graph, Dai2017relation, dai2017detecting, Li_2017_ICCV}.
3D Scene Graphs have been recently proposed as a hierarchical model of 3D environments.
Armeni \etal \citet{Armeni19iccv-3DsceneGraphs} propose 3D scene graphs as a way to capture a hierarchy of spatial entities and their
relationships in static 3D space.
Rosinol \etal \citet{Rosinol20rss-dynamicSceneGraphs,Rosinol21arxiv-Kimera} extend 3D scene graphs to represent both static space and dynamic agents, while also describing the
topology of the environment
as a graph of places.
3D scene graphs have proven useful for planning and decision-making in robotics. To this end, 
Kurenkov \etal \citet{kurenkov2021icra-HMS} propose Hierarchical Mechanical Search (HMS) to locate objects in 3D scene graphs. 
Nguyen \etal \citet{NguyenCoRL2020} learn scene graph representations from visual inputs in an unsupervised manner which are used for grasp planning.
These works use scene graphs built offline to predict high-level actions (\eg subgoals for a waypoint planner).
In contrast, we learn a policy that maps dynamically constructed scene graphs directly to low-level actions (\eg move forward, turn).


\section{Learning Navigation Policies on 3D Scene Graphs}
\label{sec:contribution}

We propose an approach to learn navigation policies for a robot that uses a 3D Dynamic Scene Graph (DSG) 
as a representation of the environment. 
Our robot is equipped with a stereo camera and an Inertial Measurement Unit.
The robot uses a Spatial PerceptIon eNgine (SPIN) similar to the one presented in~\cite{Rosinol20rss-dynamicSceneGraphs,Rosinol21arxiv-Kimera} to construct a DSG from sensor data, and
 is capable of executing a finite set of motion primitives. 
 In particular, in this paper we assume the action space $\mathcal{A} = \{\texttt{move forward}, \texttt{turn left}, \texttt{turn right}\}$, 
 but the approach can be adapted to support other action spaces including continuous ones.

Our goal is to learn a policy $\pi_\theta(a | \calG)$  with learned parameters $\theta$ that maximizes reward, where $\calG$ is a graph observation constructed directly from the DSG and $a$ is a distribution over the action space $\mathcal{A}$. 
The following sections discuss the DSG construction (Section~\ref{sec:dsg}), the graph observation $\calG$ (Section~\ref{sec:asg}), 
and the graph neural network that encodes our navigation policy (Section~\ref{sec:policy-network}).

\vspace{-0.5mm}
\subsection{3D Dynamic Scene Graph}
\label{sec:dsg}



A 3D Dynamic Scene Graph (DSG)~\cite{Rosinol20rss-dynamicSceneGraphs} represents the environment as a graph, where
nodes correspond to spatial concepts (\eg objects, places) -- potentially with a set of attributes, while edges represent relations
among concepts. The model is hierarchical, since nodes are grouped into layers
corresponding to different levels of abstraction (Fig.~\ref{fig:graph-obs}a).

Rosinol \etal \citet{Rosinol20rss-dynamicSceneGraphs} propose a DSG with five layers to describe an indoor environment. We review each layer and the corresponding nodes
and edges below. We use a subset of relevant nodes and edges to build the graph observation for our navigation policy in Section~\ref{sec:asg}.

\myParagraph{Layer 1: Metric-Semantic 3D Mesh} The lowest layer in the DSG is a dense 3D mesh, where
each Mesh node is assigned a position and semantic label, while edges describe the topology of the mesh.
In the implementation of~\cite{Rosinol20rss-dynamicSceneGraphs}, an ESDF (Euclidean Signed Distance Function) is also used to provide an alternative voxel-based representation of the
geometry of the environment in this layer.

\myParagraph{Layer 2: Objects and Agents} This layer contains both dependently mobile entities (\emph{objects}) and independently mobile entities (\emph{agents}). 
An agent is represented as a collection of nodes describing the 3D position of the agent at time $t$ and other attributes (\eg the 3D shape of the agent). The set of agents includes the robot that is  building the DSG and its trajectory. All objects have a collection of spatial attributes (\eg~3D position, bounding box) and semantic attributes (\eg~class). Each Object node is connected to the Mesh nodes in Layer 1 that comprise the object.

\myParagraph{Layer 3: Places and Structure} This layer, like {Layer 2}, is divided into two categories: places and structure. Place nodes represent free-space locations in the environment and edges between Place nodes denote straight-line traversability.
Place nodes and the corresponding edges form a topological map of the environment that can be used for path planning.
In addition, each Place node has an edge connecting it to the nearest objects and agents in {Layer 2}.
The second category of nodes (structure) represents structural elements in the environment such as walls, windows, and columns: semantically relevant regions of occupied-space in the environment.

\myParagraph{Layer 4: Rooms} Each node in Layer 4 represents a room in an indoor environment and has
a 3D location and bounding box as attributes.
Each room is also assumed to share an edge to any Place node in {Layer 3} that is contained within it, and any structural element that either comprises the room (\eg~walls) or is contained within it (\eg~columns).
Intuitively, rooms induce a grouping of Place nodes that are spatially (and likely semantically) related.
Room nodes capture both traditional concepts of a room (\eg~kitchen, bedroom, living room)
and other regions of free-space (\eg~hallways, corridors).

\myParagraph{Layer 5: Buildings} A building is the highest level of abstraction in the DSG hierarchy considered in~\cite{Rosinol20rss-dynamicSceneGraphs}.
A Building node is assumed to share an edge with any room contained within the building. 

As emphasized in~\cite{Rosinol20rss-dynamicSceneGraphs}, the DSG can be easily customized for other (indoor or outdoor) environments.
For instance,
we may add a new layer between {Layer 4} and {Layer 5} to capture the concept of \emph{floors} in a building,
or we can add additional layers at the top to model a neighborhood or a city.
Moreover, the works~\cite{Rosinol20rss-dynamicSceneGraphs, Rosinol21arxiv-Kimera} provide an automated way of building DSGs from sensor data collected by a robot, making them a suitable high-level representation to be used as an observation for a policy network.
As described in Section~\ref{sec:experiments}, we first build the overall DSG for an environment, and then selectively disclose it to the robot during training; this avoids the 
cost of rebuilding the DSG from scratch in each training episode.
We use the notation \DSGt to denote the portion of the DSG observed by the robot from the beginning of the episode until time $t$.

\vspace{-0.5mm}
\subsection{Graph Observation}
\label{sec:asg}


This section describes how we turn a DSG (including both the graph structure and the node attributes) into
a feature-space for our policy network.
This process results in a \emph{graph observation}, which
is a graph structure derived from the DSG that summarizes the environment, the robot's trajectory, and its action space.
At a high level, at each time $t$, we construct the \emph{graph observation} by first selecting a subset of nodes in
\DSGt (the portion of the scene graph observed until time $t$).
Next, we transform those nodes' positions into the robot's local reference frame.
Finally, we augment the observation with a subgraph, termed the \emph{Action layer}, which captures the robot's action space.

More formally, the graph observation at time $t$ is a pair $\graphObservationAtt = (\nodeObservationAtt, \edgeObservationAtt)$ comprising nodes $\nodeObservationAtt$ and edges $\edgeObservationAtt$.
Both a node $v^t_i \in \nodeObservationAtt$ and an edge $e^t_{ij} \in \edgeObservationAtt$
are derived from either the current scene graph \DSGt or the Action layer.
Each node $v^t_i$ has an associated feature vector $y^t_i$.
Fig.~\ref{fig:graph-obs}c shows an example graph observation.

\myParagraph{Observation Graph Structure}
To construct the observation, we first initialize the observation graph $\graphObservationAtt$ to include
all nodes and edges in \DSGt except the 3D mesh in Layer 1; this choice is motivated by computational reasons (the 3D mesh typically includes millions of vertices) and comes with little loss since the Place nodes already encode traversability.
Since we perform experiments in a single-building environment, we also omit the single node in Layer 5.
Then, we augment $\graphObservationAtt$ with a second graph $\GAL{t} = (\nodesAL{t}, \edgesAL{t})$ that we name the \emph{Action layer}, by adding nodes and edges $\nodesAL{t}$ and $\edgesAL{t}$ to $\nodeObservationAtt$ and $\edgeObservationAtt$.
Nodes $v^t_{i} \in \nodesAL{t}$ are evenly placed on a circle of radius $r_{a}$ around the agent; these represent immediate space around the robot.
We add an edge $e^t_{ij}$ to $\edgesAL{t}$ between an action node $v^t_{i} \in \nodesAL{t}$ and DSG node $v^t_{j} \in \nodeObservationAtt$
if the following three conditions are met.
First, node $v^t_{i}$ must lie in an obstacle-free location. 
To efficiently check this condition, we use the ESDF in Layer~1 of the DSG.
To determine if a given node $v^t_{i}$ lies in free-space, we simply find the value of the ESDF voxel containing the position of $v^t_{i}$
and check if the value---which describes the distance to the closest obstacle---is larger than a threshold $d_f$.
Second, $v^t_{j}$ must be within $d_{a}$ meters of $v^t_{i}$.
Finally, the straight line between $v^t_{i}$ and $v^t_{j}$ must lie within free-space. 
To determine if the straight line between two nodes $v^t_{i}$ and $v^t_{j}$ lies in free-space, we use an approximate raycast approach based on Bresenham's Line Algorithm~\citep{bresenham1965algorithm};
this algorithm retrieves the set of voxels along the line $l_{ij}$, all of which must lie in free-space for $l_{ij}$ to be considered free-space.
The resulting graph $\graphObservationAtt$ includes Place, Object, Room, and Action nodes, and the corresponding edges
(Fig.~\ref{fig:graph-obs}c). The additional graph $\GAL{t}$ essentially creates a bridge between the scene graph and the
low-level action space of the robot.

\myParagraph{Node Features}
Each node $v^t_i$ in $\nodeObservationAtt$ is assigned a 10-dimensional feature vector $y^t_i$ that captures information from both the DSG and robot's trajectory history.
The first three elements of $y^t_i$ contain the position of node $v^t_i$ relative to the robot.
The next three elements encode the dimensions of an object's bounding box, if the node is associated with an object; otherwise, these indices are assigned a \texttt{NULL} value.
The next two elements indicate the node category (\eg object, place, room, action) and, if applicable, its semantic class (\eg office, hallway, desk, chair, etc.).
The next entry is binary and describes information relevant to robot traversal.
In particular, this entry is equal to 1 under two conditions: (1) the node belongs to the DSG and the robot's trajectory has come within $d_t$ meters of the node's location, or (2) the node is in the Action layer and lies in free-space; otherwise, this value is 0.
Finally, the last element contains $v^t_i$'s ESDF value if $v^t_i$ is a Place or Action node, 0 if $v^t_i$ is an Object node, and a \texttt{NULL} value if $v^t_i$ is a Room node.

While in Section~\ref{sec:experiments} we observe this choice of features to produce good results,
we remark that one may also adopt more complex
feature embeddings (\eg encode semantic classes using vector embeddings as in~\cite{Sunderhauf2020arxiv-keys} or add extra dimensions to
capture other attributes of the nodes).

\subsection{Graph Neural Network Policy}
\label{sec:policy-network}


Given a graph observation $\graphObservationAtt$, 
we perform message passing on $\graphObservationAtt$ using a standard GNN to extract a single, graph-level feature vector.
This feature vector is then used by a policy network to predict a distribution over actions. We discuss each step below.

\myParagraph{Graph Neural Network}
\label{sec:gnns}
We compute a feature vector for the entire graph using message passing on $\graphObservationAtt$.
At each iteration $k$ of message passing, each node's feature vector ${y}^{t,k}_{i}, \forall v^t_i \in \nodeObservationAtt$ is updated by aggregating the feature vectors of the node's neighbors $\mathcal{N}(v^t_i)$ in graph $\graphObservationAtt$:
\begin{equation}
  \label{eq:GNN_AGG}
  y^{t,k}_{i} = AGG_{k}\left( y^{t,k-1}_{i}, \{ y^{t,k-1}_{j}~|~v^t_j \in \mathcal{N}(v^t_i) \}\right).
\end{equation}
The initial feature vector for each node is set to ${y}^{t,0}_{i} = y^{t}_i, \forall v^{t}_i \in \nodeObservationAtt$,
where $y^{t}_i$ are the node features in the graph observation described in Section~\ref{sec:asg}.
Related works describe difference choices of aggregation functions (\eg~GCN~\cite{kipf2016semi}, GAT~\cite{Velickovic18iclrGAT}, GraphSAGE~\cite{hamilton2017GarphSage}); 
in this work, we use a 64-channel Graph Convolutional Operator~\cite{kipf2016semi} 
followed by a ReLU activation function. 
After $K$ iterations of message passing ($K=3$ in our implementation), 
a graph-level feature vector is extracted as
\begin{equation}\label{eq:read}
  y^t_{\mathcal{G}} = READ_{\mathcal{G}}\left( \{ y^{t, K}_{v}~|~v \in \mathcal{V}\}\right)\,,
\end{equation}
where $\{ y^{t, K}_{v}~|~v \in \mathcal{V}\}$ denotes a tuple of node feature vectors at iteration $K$, 
and $\mathcal{V}$ is a subset of the nodes $\nodeObservationAtt$  that captures relevant information about the navigation task. 
In our implementation, 
we define $READ_{\calG}$ as a Multilayer Perceptron (MLP) with two layers of length 512 that takes as input the stacked Action layer feature vectors $[y^{t,K}_{a_0}, y^{t,K}_{a_1}, ...]$, \ie we use $\calV = \mathcal{V}^t_a$ in~\eqref{eq:read}.



\myParagraph{Policy Architecture}
\label{sec:policy_architecture}
We use an Actor-Critic method~\cite{Konda00actorcriticalgorithms}, where an actor (\ie policy) $\pi_\theta(a^t | \calG^t)$, parameterized by $\theta$, predicts action distribution $a^t$ given observation $\calG^t$ and critic $V^\pi_\phi(\calG^t)$, parameterized by $\phi$, predicts the expected return given $\calG^t$.
Actor $\pi_\theta(a^t | \calG^t)$ and critic $V^\pi_\phi(\calG^t)$ use an identical backbone function $f(\calG^t)$, such that $\pi_\theta(a^t | \calG^t) = \pi_\theta(a^t | f(\calG^t))$ and $V^\pi_\phi(\calG^t) = V^\pi_\phi(f(\calG^t))$.
We implement the backbone function using the GNN described above.
The resulting feature vector $y^t_{\mathcal{G}}= f(\calG^t)$ is then used by the Actor and Critic networks, each of which are parameterized by an 1-layer MLP mapping $y^t_{\mathcal{G}}$ to the action distribution and return estimate, respectively. 
We train our policy using Proximal Policy Optimization~\citep{schulman2017ppo} for 10$^6$ steps. 


\section{Experiments}
\label{sec:experiments}

We consider a multi-object search task with a robot exploring a previously unseen multi-room office environment.
Multiple targets are only placed within \emph{target} rooms, which are semantically distinct from \emph{distractor} rooms.
Target rooms correspond to offices and conference rooms, while distractor rooms are hallways, bathrooms, breakrooms, and storage rooms.
The robot must find as many targets as possible.
An effective policy 
\highlight{must} reason about both local navigation to avoid obstacles 
\emph{and} reason about high-level navigation to focus search 
\highlight{on target rooms,}
while also limiting the amount of area revisited.
This section describes the experimental setup (Section~\ref{sec:setup}),
shows that the proposed approach outperforms popular baselines (Section~\ref{sec:baseline-comparison}),
and provides ablation studies to assess the importance of hierarchical representations and explicit memory (Section~\ref{sec:hierarchy-and-memory})
and the role of the Action layer (Section~\ref{sec:robot-ablation}).

\subsection{Experimental Setup}
\label{sec:setup}

\myParagraph{Simulation Environment} We conduct experiments with a photo-realistic Unity-based simulator of indoor office buildings that includes a diverse set of assets, providing visual and spatial variations across scenes.
We consider seven environments with a cumulative area of 8,730\,m$^2$, which are split into train and test sets with areas of 3,880\,m$^2$ and 4,850\,m$^2$, respectively.
Targets are invisible objects \highlight{in \textit{target} rooms} that the agent automatically collects when within 2\,m range.
Once a target is collected, it is removed from the scene and the agent receives a reward of 1.
At each step, the agent may move forward 0.5\,m, turn left 8 degrees, or turn right 8 degrees.
We randomly place 30 targets in the scene and set the maximum number of steps to 400 for both train and test episodes.
An episode ends after either all targets are collected or the maximum number of steps are taken.

\begin{figure}[ht!]
    \vspace{-2mm}
    \newcommand{\baselineobswidth}{0.225}
    \centering
    \subfloat[RGB]{
        \includegraphics[width=\baselineobswidth\textwidth, trim=0 60 0 5, clip]{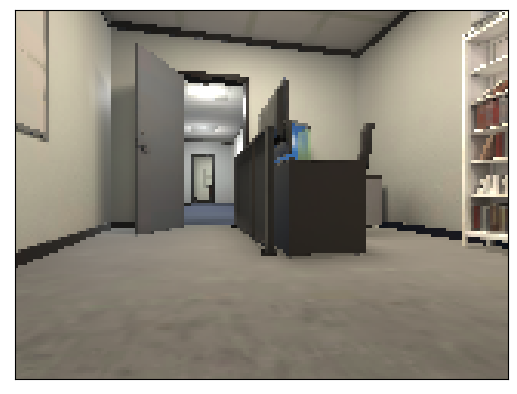}
    }
    \subfloat[2D Segmentation]{
        \includegraphics[width=\baselineobswidth\textwidth, trim=0 60 0 5, clip]{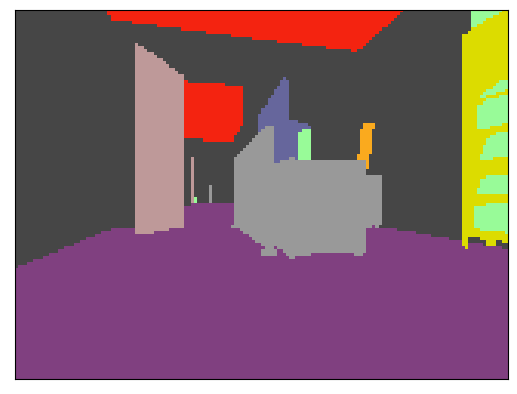}
    }
    \vspace{-0.25cm}
    \subfloat[Depth]{
        \includegraphics[width=\baselineobswidth\textwidth, trim=0 60 0 5, clip]{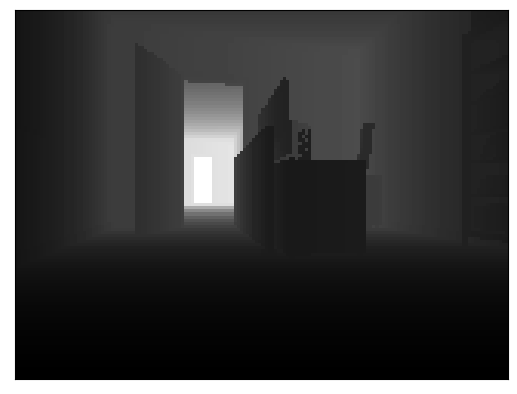}
    }
    \subfloat[ESDF Slice]{
    \includegraphics[width=\baselineobswidth\textwidth, trim=0 60 0 5, clip]{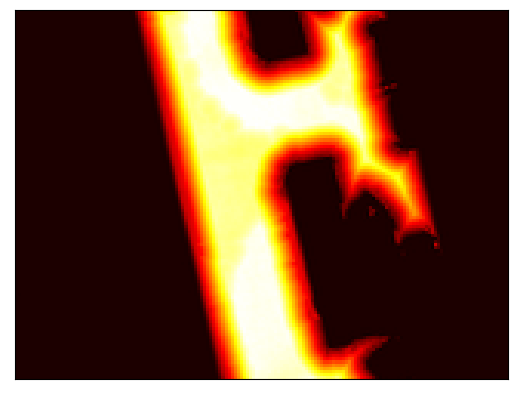}
    }
    \caption{\highlight{Example observations generated by the simulator.\label{fig:baseline_obs}\vspace{-5mm}}}
\end{figure}

The simulator provides RGB imagery, 2D semantic segmentation, depth, and inertial data as illustrated in Fig.~\ref{fig:baseline_obs}.
These ground-truth observations are used by all the approaches.
The simulator also provides an horizontal slice of the ESDF centered on the robot and aligned with its heading,
which is used by one of our baselines.

\myParagraph{DSG Creation}
Since creating a DSG of a scene at each time step and at each training episode is computationally expensive~\cite{Rosinol21arxiv-Kimera},
we first build and store a DSG of each scene offline.
Then, at runtime, we
simulate \emph{online} operation by incrementally exposing the DSG depending on the robot motion.
As the robot moves across space and observes nodes from the offline DSG, these nodes are \emph{accumulated} into the online DSG.
%
More specifically, 
at each time $t$, we construct the set $\bar{\calV}^t$ comprised of Places nodes visible from the robot's on-board camera at time $t$.
A node is considered visible if it lies within the robot's field of view and is not occluded.
We consult the ESDF to determine occlusion using the process described in Section~\ref{sec:asg}.
Any Room or Object nodes connected to a node in $\bar{\calV}^t$ are then added to the set.
Finally, we add this set of ``newly observed'' nodes and the corresponding edges to the DSG built at time $t-1$ (this graph is empty at $t=0$). After the online DSG at time $t$ is built, we obtain the graph observation as discussed in Section~\ref{sec:asg}.
We set the free-space and traversal thresholds $d_f$ and $d_t$ to 0.1\,m and 2\,m.
The Action layer uses 8 nodes with a radius $r_{a}$ and edge threshold $d_{a}$ of 1\,m.
Figure~\ref{fig:accumulation} illustrates example DSGs built over the course of an episode.

\begin{figure}[ht]
    \begin{minipage}{0.25\textwidth}
            \centering
            \scalebox{4.}{\input{includes/graph_accumulation_t1.tex}}
            \caption*{(a) t=1}
            \vspace{0.1cm}
            \scalebox{3.5}{\input{includes/graph_accumulation_t100.tex}}
            \caption*{(b) t=100}
    \end{minipage}%
    \begin{minipage}{0.25\textwidth}
            \centering
            \scalebox{4.1}{\input{includes/graph_accumulation_t400.tex}}
            \caption*{(c) t=400}
    \end{minipage}
    \caption{\highlight{Top-down view of a DSG during an episode.
        Each subfigure shows the cumulative DSG observed up to time $t$.\label{fig:accumulation}\vspace{-4mm}}}
\end{figure}

\subsection{Comparison with Baselines}
\label{sec:baseline-comparison}

We compare our method against three policies that use the following representations: RGB, depth, semantic segmentation, and ESDFs.
The first policy takes an RGB image as its observation (label: ``RGB'').
The second takes RGB, depth, and semantic segmentation imagery (``RGB-D + Sem.'').
The third takes a top-down view of the scene's ESDF centered around the robot (``ESDF'').
All three policies use a CNN modeled after AtariNet~\cite{Mnih2015Nature} to extract features from visual inputs.
The resulting feature vector is given to a length-512 linear layer, concatenated with the agent's ground-truth pose, then given to a Recurrent Neural Network (RNN).
The RNN output is used by actor and critic networks, which each consist of a linear layer.
All baselines are trained with Proximal Policy Optimization~\cite{schulman2017ppo} for 10$^6$ steps.

We report three evaluation metrics: percentage of targets found, number of collisions, and area explored.
We calculate area explored by discretizing the environment into 1\,m$^2$ cells and tallying the number of visited cells.

Table~\ref{tab:main_results} shows \highlight{that}
our method outperforms all baselines in \highlight{in terms of} number of found targets and area explored,
\highlight{which indicates improved semantic understanding and memory retention}.
\highlight{RGB-D + Sem. and RGB induce less collisions than our method; 
this performance gap may be due to the sparse nature of our method's representations.
Visual inputs provide dense occupancy information about the robot's immediate free-space, 
whereas our method must infer free-space from the sparse Places layer.
}
Surprisingly, ESDF has the most collisions. 
Qualitative analysis suggests that this policy learns navigation strategies that are prone to colliding with obstacle edges (\eg doorways, table corners).

\begin{table}[ht]
    \centering
    {\scriptsize
    \begin{tabular}{llll}
        \toprule
        \multirow{ 2}{*}{Method}            & Targets Found & Collisions  & Area Explored \\
                     & (\%, $\uparrow$) & ($\downarrow$)  & (m$^2$, $\uparrow$) \\
        \midrule
        Proposed          & \textbf{44.2} (42.7, 45.7)     & 90.0 (84.3, 95.6)          & \textbf{59.1} (57.7, 60.5)        \\
        RGB-D + Sem.        & 39.7 (38.4, 40.8)              & \textbf{45.3} (41.7, 48.8) & 53.3 (52.4, 54.3)                 \\
        RGB               & 38.4 (37.4, 39.4)              & 82.4 (78.2, 86.3)          & 50.0 (48.9, 51.1)                 \\
        ESDF              & 31.6 (30.3, 32.9)              & 197.5 (190.7, 204.2)        & 38.6 (37.3, 39.8) \\
        \bottomrule
    \end{tabular}
    }
    \caption{Results with 95\% confidence intervals.\label{tab:main_results}\vspace{-5mm}} 
\end{table}

\subsection{Hierarchical Representations and Explicit Memory}
\label{sec:hierarchy-and-memory}

Next, we evaluate the impact of hierarchical information and explicit memory on our approach.
To evaluate the contribution of hierarchical information, we run our method as described in Section~\ref{sec:contribution}, but only use the Places layer of the DSG.
To evaluate the contribution of using explicit memory, we remove all notion of history from our graphical representations by skipping DSG accumulation.
That is, we compare against a version of the proposed approach where the observation $\mathcal{G}^t$ is formed
only from the portion of the DSG directly observed at time $t$, and we also remove the binary indicator in each node feature vector
that describes whether that node has been visited in the past.

The results in Table~\ref{tab:dsg_ablation} show that hierarchical information does indeed improve the number of targets found.
Interestingly, the improved performance is not the result of fewer collisions or exploring more area; instead, the agent is able to focus its search on areas where targets are most likely. 
The second key finding is that retaining explicit memory in the DSG is important.
Even though a robot with ``no memory'' is better at avoiding collisions (last row in Table~\ref{tab:dsg_ablation}), it also explores less area indicating that it revisits areas more often compared to the proposed approach.
\begin{table}[ht]
    \vspace{3mm}
    \centering
    {\scriptsize
    \begin{tabular}{llll}
        \toprule
        \multirow{ 2}{*}{Method}            & Targets Found & Collisions  & Area Explored \\
                     & (\%, $\uparrow$) & ($\downarrow$)  & (m$^2$, $\uparrow$) \\
        \midrule
        Proposed     & \textbf{44.2} (42.7, 45.7)     & 90.0 (84.3, 95.6)          & \textbf{59.0} (57.7, 60.5)        \\
        No Hierarchy & 41.9 (40.5, 43.3)              & 91.9 (85.6, 98.3)          & 58.3 (56.8, 59.7)                 \\
        No Memory    & 38.6 (37.3, 39.9)              & \textbf{67.2} (61.6, 72.4) & 53.0 (51.6, 54.4)                 \\
        \bottomrule
    \end{tabular}
    }
    \caption{Ablation: Importance of hierarchical representations and explicit memory. \label{tab:dsg_ablation}\vspace{-5mm}}
\end{table}

\subsection{The Role of the Action Layer}
\label{sec:robot-ablation}

The Action layer is a key aspect of the proposed architecture.
In this section we assess the importance of adding this layer and our design choice of how to connect the Action layer with the DSG.
First, we consider the effect of removing the occlusion-checking procedure described in Section~\ref{sec:asg}.
For this experiment, we do not check for free-space when connecting Action layer nodes to DSG nodes.
Thus, a node in the Action layer could be placed in occupied space; or, an edge could penetrate an obstacle.
Our second experiment considers the overall importance of the Action layer by completely removing it.
We use a policy nearly identical to that described in Section~\ref{sec:policy-network}.
However, now the graph observation does not include the Action layer and 
the graph-level vector $y^t_{\mathcal{G}}$ in~\eqref{eq:read}  is computed from the 
nodes in the DSG at time $t$ rather than the Action nodes.

The results of these experiments, shown in Table~\ref{tab:agent_layer_ablation}, demonstrate that both the Action layer (``No AL'') and the occlusion-checking procedure (``No Occ.'') we use to connect the Action layer to the DSG are crucial to the performance of our method.
Incorporating occlusion checking reduces collisions by more than half, leading to an improvement in targets found and area explored.
Without the Action layer, the robot is unable to learn mappings from the graphical scene representation to its low-level action space.
\begin{table}[ht]
    \centering
    {\scriptsize
    \begin{tabular}{p{1.5cm}lll}
        \toprule
        \multirow{ 2}{*}{Method}            & Targets Found & Collisions  & Area Explored \\
                     & (\%, $\uparrow$) & ($\downarrow$)  & (m$^2$, $\uparrow$) \\
        \midrule
        Proposed              & \textbf{44.2} (42.7, 45.2)     & \textbf{90.0} (84.3, 95.6) & \textbf{59.1} (57.7, 60.5)        \\
        No Occ. Check &  31.7 (30.3, 33.0)            & 229.8 (222.6, 236.9)      & 35.7 (34.3, 37.1) \\
        No  AL     & 8.7 (8.2, 9.2)                & 377.9 (374.2, 382.0)      & 8.5 (7.8, 9.2) \\
        \bottomrule
    \end{tabular}
    }
    \caption{Ablation: Role and design of the Action layer. \label{tab:agent_layer_ablation} \vspace{-5mm}}
\end{table}


\section{Conclusion}
\label{sec:conclusion}

We proposed an approach to learn effective robot navigation policies using a 3D scene graph (in particular, a DSG).
Our experiments show the advantages of explicit memory and hierarchical representations encoded in the DSG for learning high-quality policies
compared to more typical visuomotor policies.
\highlight{Obstacle avoidance}
using solely graphical data remains a challenge; 
future work could further refine the Action Layer or incorporate dense depth sources (\eg from the DSG's Mesh layer).
We believe there are several other avenues for future work including developing more complex node feature
embeddings and investigating other GNN message passing schemes.
\highlight{
We are especially interested in integrating Hydra~\cite{Hughes2022Hydra}, a method for real-time DSG construction, 
which was proposed in parallel to our work and would enable experiments on physical robots.
}




\clearpage



\bibliographystyle{IEEEtran}
\bibliography{./includes/refs.bib,./includes/myRefs.bib,./includes/rl_abstractions_refs.bib,./includes_supplemental/supplemental_refs.bib}

\begin{thebibliography}{10}
\providecommand{\url}[1]{#1}
\csname url@rmstyle\endcsname
\providecommand{\newblock}{\relax}
\providecommand{\bibinfo}[2]{#2}
\providecommand\BIBentrySTDinterwordspacing{\spaceskip=0pt\relax}
\providecommand\BIBentryALTinterwordstretchfactor{4}
\providecommand\BIBentryALTinterwordspacing{\spaceskip=\fontdimen2\font plus
\BIBentryALTinterwordstretchfactor\fontdimen3\font minus
  \fontdimen4\font\relax}
\providecommand\BIBforeignlanguage[2]{{%
\expandafter\ifx\csname l@#1\endcsname\relax
\typeout{** WARNING: IEEEtran.bst: No hyphenation pattern has been}%
\typeout{** loaded for the language `#1'. Using the pattern for}%
\typeout{** the default language instead.}%
\else
\language=\csname l@#1\endcsname
\fi
#2}}

\bibitem{li2017deep}
Y.~Li, ``Deep reinforcement learning: An overview,'' \emph{arXiv preprint
  arXiv:1701.07274}, 2017.

\bibitem{mnih2013playing}
V.~Mnih, K.~Kavukcuoglu, D.~Silver, A.~Graves, I.~Antonoglou, D.~Wierstra, and
  M.~Riedmiller, ``Playing atari with deep reinforcement learning,''
  \emph{arXiv preprint arXiv:1312.5602}, 2013.

\bibitem{Sax2019corl-MidLevelReps}
A.~Sax, B.~Emi, A.~Zamir, L.~Guibas, S.~Savarese, and J.~Malik, ``Mid-level
  representations improve generalization and sample efficiency for learning
  visuomotor policies,'' in \emph{Conference on Robot Learning (CoRL)}, 2019.

\bibitem{Chen2020corl-MidLevelReps}
B.~Chen, A.~Sax, G.~Lewis, I.~Armeni, S.~Savarese, A.~Zamir, J.~Malik, and
  L.~Pinto, ``Robust policies via mid-level visual representations: An
  experimental study in manipulation and navigation,'' in \emph{Conference on
  Robot Learning (CoRL)}, 2020.

\bibitem{Zhou2019ScienceRobotics}
B.~Zhou, P.~Krahenbuhl, and V.~Koltun, ``Does computer vision matter for
  action,'' \emph{Science Robotics}, 2019.

\bibitem{Muller2018Corl}
M.~Muller, A.~Dosovitskiy, B.~Ghanem, and V.~Koltun, ``Driving policy transfer
  via modularity and abstraction,'' in \emph{Conference on Robot Learning
  (CoRL)}, 2018.

\bibitem{Mousavian2019ICRA}
A.~Mousavian, A.~Toshev, M.~Fiser, J.~Kosecka, A.~Wahid, and J.~Davidson,
  ``Visual representations for semantic target driven navigation,'' in
  \emph{IEEE Intl. Conf. on Robotics and Automation (ICRA)}, 2019.

\bibitem{wu2016training}
Y.~Wu and Y.~Tian, ``Training agent for first-person shooter game with
  actor-critic curriculum learning,'' in \emph{International Conference on
  Learning Representations (ICLR)}, 2017.

\bibitem{shehroze2016doom}
S.~Bhatti, A.~Desmaison, O.~Miksik, N.~Nardelli, N.~Siddharth, and P.~H.~S.
  Torr, ``Playing doom with slam-augmented deep reinforcement learning.''
  \emph{arXiv preprint arxiv:1612.00380}, 2016.

\bibitem{zhang2017neural}
J.~Zhang, L.~Tai, J.~Boedecker, W.~Burgard, and M.~Liu, ``Neural slam: Learning
  to explore with external memory,'' \emph{arXiv preprint arXiv:1706.09520},
  2017.

\bibitem{Chaplot2020CVPR}
D.~Chaplot, R.~Salakhutdinov, A.~Gupta, and S.~Gupta, ``Neural topological slam
  for visual navigation,'' in \emph{IEEE Conf. on Computer Vision and Pattern
  Recognition (CVPR)}, 2020.

\bibitem{Tateno17cvpr-CNN-SLAM}
K.~{Tateno}, F.~{Tombari}, I.~Laina, and N.~{Navab}, ``{CNN-SLAM}: Real-time
  dense monocular slam with learned depth prediction,'' in \emph{IEEE Conf. on
  Computer Vision and Pattern Recognition (CVPR)}, 2017.

\bibitem{Lianos18eccv-VSO}
K.-N. Lianos, J.~L. Sch{\"o}nberger, M.~Pollefeys, and T.~Sattler, ``Vso:
  Visual semantic odometry,'' in \emph{European Conf. on Computer Vision
  (ECCV)}, 2018, pp. 246--263.

\bibitem{Dong17cvpr-XVIO}
J.~Dong, X.~Fei, and S.~Soatto, ``Visual-inertial-semantic scene representation
  for {3D} object detection,'' in \emph{IEEE Conf. on Computer Vision and
  Pattern Recognition (CVPR)}, 2017.

\bibitem{Behley19iccv-semanticKitti}
J.~Behley, M.~Garbade, A.~Milioto, J.~Quenzel, S.~Behnke, C.~Stachniss, and
  J.~Gall, ``{SemanticKITTI: A Dataset for Semantic Scene Understanding of
  LiDAR Sequences},'' in \emph{Intl. Conf. on Computer Vision (ICCV)}, 2019.

\bibitem{McCormac17icra-semanticFusion}
J.~McCormac, A.~Handa, A.~J. Davison, and S.~Leutenegger, ``{SemanticFusion:
  Dense 3D Semantic Mapping with Convolutional Neural Networks},'' in
  \emph{IEEE Intl. Conf. on Robotics and Automation (ICRA)}, 2017.

\bibitem{Zheng19arxiv-metricSemantic}
L.~Zheng, C.~Zhu, J.~Zhang, H.~Zhao, H.~Huang, M.~Niessner, and K.~Xu, ``Active
  scene understanding via online semantic reconstruction,'' \emph{arXiv
  preprint:1906.07409}, 2019.

\bibitem{Tateno15iros-metricSemantic}
K.~{Tateno}, F.~{Tombari}, and N.~{Navab}, ``Real-time and scalable incremental
  segmentation on dense slam,'' in \emph{IEEE/RSJ Intl. Conf. on Intelligent
  Robots and Systems (IROS)}, 2015, pp. 4465--4472.

\bibitem{Li16iros-metricSemantic}
C.~{Li}, H.~{Xiao}, K.~{Tateno}, F.~{Tombari}, N.~{Navab}, and G.~D. {Hager},
  ``Incremental scene understanding on dense {SLAM},'' in \emph{IEEE/RSJ Intl.
  Conf. on Intelligent Robots and Systems (IROS)}, 2016, pp. 574--581.

\bibitem{McCormac183dv-fusion++}
J.~McCormac, R.~Clark, M.~Bloesch, A.~J. Davison, and S.~Leutenegger,
  ``Fusion++: Volumetric object-level {SLAM},'' in \emph{Intl. Conf. on 3D
  Vision (3DV)}, 2018, pp. 32--41.

\bibitem{Runz18ismar-maskfusion}
M.~Runz, M.~Buffier, and L.~Agapito, ``Maskfusion: Real-time recognition,
  tracking and reconstruction of multiple moving objects,'' in \emph{IEEE
  International Symposium on Mixed and Augmented Reality (ISMAR)}.\hskip 1em
  plus 0.5em minus 0.4em\relax IEEE, 2018, pp. 10--20.

\bibitem{Runz17icra-cofusion}
M.~R{\"u}nz and L.~Agapito, ``Co-fusion: Real-time segmentation, tracking and
  fusion of multiple objects,'' in \emph{IEEE Intl. Conf. on Robotics and
  Automation (ICRA)}.\hskip 1em plus 0.5em minus 0.4em\relax IEEE, 2017, pp.
  4471--4478.

\bibitem{Xu19icra-midFusion}
B.~Xu, W.~Li, D.~Tzoumanikas, M.~Bloesch, A.~Davison, and S.~Leutenegger,
  ``{MID-Fusion}: Octree-based object-level multi-instance dynamic slam,'' in
  \emph{IEEE Intl. Conf. on Robotics and Automation (ICRA)}, 2019, pp.
  5231--5237.

\bibitem{Rosinol20icra-Kimera}
A.~Rosinol, M.~Abate, Y.~Chang, and L.~Carlone, ``Kimera: an open-source
  library for real-time metric-semantic localization and mapping,'' in
  \emph{IEEE Intl. Conf. on Robotics and Automation (ICRA)}, 2020.

\bibitem{Grinvald19ral-voxbloxpp}
M.~{Grinvald}, F.~{Furrer}, T.~{Novkovic}, J.~J. {Chung}, C.~{Cadena},
  R.~{Siegwart}, and J.~{Nieto}, ``{Volumetric Instance-Aware Semantic Mapping
  and 3D Object Discovery},'' \emph{{IEEE} Robotics and Automation Letters},
  vol.~4, no.~3, pp. 3037--3044, 2019.

\bibitem{Rosinol20rss-dynamicSceneGraphs}
A.~Rosinol, A.~Gupta, M.~Abate, J.~Shi, and L.~Carlone, ``{3D} dynamic scene
  graphs: Actionable spatial perception with places, objects, and humans,'' in
  \emph{Robotics: Science and Systems (RSS)}, 2020.

\bibitem{Armeni19iccv-3DsceneGraphs}
I.~Armeni, Z.-Y. He, J.~Gwak, A.~R. Zamir, M.~Fischer, J.~Malik, and
  S.~Savarese, ``{3D} scene graph: A structure for unified semantics, {3D}
  space, and camera,'' in \emph{Intl. Conf. on Computer Vision (ICCV)}, 2019,
  pp. 5664--5673.

\bibitem{Rosinol21arxiv-Kimera}
A.~Rosinol, A.~Violette, M.~Abate, N.~Hughes, Y.~Chang, J.~Shi, A.~Gupta, and
  L.~Carlone, ``Kimera: from {SLAM} to spatial perception with {3D} dynamic
  scene graphs,'' \emph{arXiv preprint arXiv: 2101.06894}, 2021.

\bibitem{kurenkov2021icra-HMS}
A.~Kurenkov, R.~Mart{\'\i}n-Mart{\'\i}n, J.~Ichnowski, K.~Goldberg, and
  S.~Savarese, ``Semantic and geometric modeling with neural message passing in
  3d scene graphs for hierarchical mechanical search,'' \emph{arXiv preprint
  arXiv:2008.07792}, 2020.

\bibitem{Sunderhauf2020arxiv-keys}
N.~Sunderhauf, ``Where are the keys? - learning object-centric navigation
  policies on semantic maps with graph convolutional networks,'' \emph{arXiv
  preprint arXiv:1909.07376}, 2020.

\bibitem{ZhuCoRL2020}
\BIBentryALTinterwordspacing
Y.~Zhu, J.~Tremblay, S.~Birchfield, and Y.~Zhu, ``Hierarchical planning for
  long-horizon manipulation with geometric and symbolic scene graphs,'' 2020.
  [Online]. Available:
  \url{https://zhuyifengzju.github.io/projects/hierarchical-scene-graph/}
\BIBentrySTDinterwordspacing

\bibitem{NguyenCoRL2020}
\BIBentryALTinterwordspacing
S.~Nguyen, O.~Oguz, V.~Hartmann, and M.~Toussaint, ``Self-supervised learning
  of scene-graph representations to solve sequential manipulation problems,''
  in \emph{Conference on Robot Learning (CoRL)}, 2020. [Online]. Available:
  \url{https://github.com/sontung/location-based-generative}
\BIBentrySTDinterwordspacing

\bibitem{Fulda2017corl}
N.~Fulda, N.~Tibbetts, Z.~Brown, and D.~Wingate, ``Harvesting common-sense
  navigational knowledge for robotics from uncurated text corpora,'' in
  \emph{Conference on Robot Learning (CoRL)}, 2017.

\bibitem{thomasonaaai18}
\BIBentryALTinterwordspacing
J.~Thomason, J.~Sinapov, R.~Mooney, and P.~Stone, ``Guiding exploratory
  behaviors for multi-modal grounding of linguistic descriptions,'' in
  \emph{Proceedings of the Thirty-Second AAAI Conference on Artificial
  Intelligence (AAAI-18)}, February 2018. [Online]. Available:
  \url{http://www.cs.utexas.edu/users/ai-labpub-view.php?PubID=127682}
\BIBentrySTDinterwordspacing

\bibitem{darunaicra2019}
A.~{Daruna}, W.~{Liu}, Z.~{Kira}, and S.~{Chetnova}, ``Robocse: Robot common
  sense embedding,'' in \emph{2019 International Conference on Robotics and
  Automation (ICRA)}, 2019.

\bibitem{Yang2019icml}
W.~Yang, X.~Wang, A.~Farhadi, A.~Gupta, and R.~Mottaghi, ``Visual semantic
  navigation using scene priors,'' in \emph{Intl. Conf. on Machine Learning
  (ICML)}, 2019.

\bibitem{MuraliCoRL2020}
\BIBentryALTinterwordspacing
A.~Murali, W.~Liu, K.~Marino, S.~Chernova, and A.~Gupta, ``Same object,
  different grasps: Data and semantic knowledge for task-oriented grasping,''
  in \emph{Conference on Robot Learning (CoRL)}, 2020. [Online]. Available:
  \url{https://sites.google.com/view/taskgrasp/home}
\BIBentrySTDinterwordspacing

\bibitem{Scaliseiros19}
R.~Scalise, J.~Thomason, Y.~Bisk, and S.~S. Srinivasa, ``Improving robot
  success detection using static object data,'' in \emph{2019 {IEEE/RSJ}
  International Conference on Intelligent Robots and Systems, {IROS}}.\hskip
  1em plus 0.5em minus 0.4em\relax {IEEE}, 2019.

\bibitem{Arkinijrr2020}
J.~Arkin, D.~Park, S.~Roy, M.~R. Walter, N.~Roy, T.~M. Howard, and R.~Paul,
  ``Multimodal estimation and communication of latent semantic knowledge for
  robust execution of robot instructions,'' \emph{The International Journal of
  Robotics Research}, vol.~39, no. 10-11, pp. 1279--1304, 2020.

\bibitem{DuECCV2020}
H.~Du, X.~Yi, and L.~Zheng, ``Learning object relation graph and tentative
  policy for visual navigation,'' in \emph{European Conference on Computer
  Vision}, 2020.

\bibitem{Druon2020RAL}
R.~Druon, Y.~Yoshiyasu, A.~Kanezakiand, and A.~Watt, ``Visual object search by
  learning spatial context,'' \emph{{IEEE} Robotics and Automation Letters},
  2020.

\bibitem{chaplot2020object}
D.~S. Chaplot, D.~Gandhi, A.~Gupta, and R.~Salakhutdinov, ``Object goal
  navigation using goal-oriented semantic exploration,'' in \emph{In Neural
  Information Processing Systems (NeurIPS)}, 2020.

\bibitem{Qui2020CORL}
\BIBentryALTinterwordspacing
Y.~Qiu, A.~Pal, and H.~Christensen, ``Learning hierarchical relationships for
  object-goal navigation,'' in \emph{Conference on Robot Learning (CoRL)},
  2020. [Online]. Available:
  \url{https://sites.google.com/eng.ucsd.edu/mjolnir}
\BIBentrySTDinterwordspacing

\bibitem{Krishnaijcv2017}
O.~Ranjay~Krishna, Yuke an~Zhu, Groth, J.~Johnson, K.~Hata, J.~Kravitz,
  S.~Chen, Y.~Kalantidis, L.-J. Li, D.~A. Shamma, M.~Bernstein, and L.~Fei-Fei,
  ``Visual genome: Connecting language and vision using crowdsourced dense
  image annotations,'' \emph{arxiv:1602.07332}, 2016.

\bibitem{ChenICLR2019}
\BIBentryALTinterwordspacing
T.~Chen, S.~Gupta, and A.~Gupta, ``Learning exploration policies for
  navigation,'' in \emph{International Conference on Learning Representations
  (ICLR)}, 2019. [Online]. Available:
  \url{https://sites.google.com/view/exploration-for-nav/}
\BIBentrySTDinterwordspacing

\bibitem{MirowskiICLR2016}
P.~Mirowski, R.~Pascanu, F.~Viola, H.~Soyer, A.~Ballard, A.~Banino, M.~Denil,
  R.~Goroshin, L.~Sifre, K.~Kavukcuoglu, D.~Kumaran, and R.~Hadsell, ``Learning
  to navigate in complex environments,'' in \emph{International Conference on
  Learning Representations (ICLR)}, 2017.

\bibitem{chaplot2020learning}
D.~S. Chaplot, D.~Gandhi, S.~Gupta, A.~Gupta, and R.~Salakhutdinov, ``Learning
  to explore using active neural slam,'' in \emph{International Conference on
  Learning Representations (ICLR)}, 2020.

\bibitem{ParisottoICLR2018}
Y.~Wu, Y.~Wu, A.~Tamar, S.~Russell, G.~Gkioxari, and Y.~Tian, ``Neural map:
  Structured memory for deep reinforcement learning,'' in \emph{International
  Conference on Learning Representations (ICLR)}, 2018.

\bibitem{Gupta2020IJCV}
\BIBentryALTinterwordspacing
S.~Gupta, V.~Tolani, J.~Davidson, S.~Levine, R.~Sukthankar, and J.~Malik,
  ``Cognitive mapping and planning for visual navigation,'' \emph{Intl. J. of
  Computer Vision}, 2020. [Online]. Available:
  \url{https://sites.google.com/view/cognitive-mapping-and-planning/}
\BIBentrySTDinterwordspacing

\bibitem{Wu2019ICCV}
Y.~Wu, Y.~Wu, A.~Tamar, S.~Russell, G.~Gkioxari, and Y.~Tian, ``Bayesian
  relational memory for semantic visual navigation,'' in \emph{Intl. Conf. on
  Computer Vision (ICCV)}, 2019.

\bibitem{Chen2019RSS}
K.~Chen, J.~P. de~Vicente, G.~Sepulveda, A.~S. F.~Xia, M.~Vazquez, and
  S.~Savarese, ``A behavioral approach to visual navigation with graph
  localization networks,'' in \emph{Robotics: Science and Systems (RSS)}, 2019.

\bibitem{BeechingECCV2020}
E.~Beeching, J.~Dibangoye, O.~Simonin, and C.~Wolf, ``Learning to plan with
  uncertain topological maps,'' in \emph{European Conference on Computer
  Vision}, 2020.

\bibitem{SavinovICLR2018}
N.~Savinov, A.~Dosovitskiy, and V.~Koltun, ``Semi-parametric topological memory
  for navigation,'' in \emph{International Conference on Learning
  Representations (ICLR)}, 2018.

\bibitem{Chen2020GCNNExploration}
F.~Chen, J.~D. Martin, Y.~Huang, J.~Wang, and B.~Englot, ``Autonomous
  exploration under uncertainty via deep reinforcement learning on graphs,'' in
  \emph{IEEE/RSJ Intl. Conf. on Intelligent Robots and Systems (IROS)}, 2020.

\bibitem{Jiang2020GraphConvRL}
J.~Jiang, C.~Dun, T.~Huang, and Z.~Lu, ``Graph convolutional reinforcement
  learning,'' in \emph{International Conference on Learning Representations
  (ICLR)}, 2020.

\bibitem{hu2020GraphPolicyNetwork}
S.~Hu, Z.~Xiong, M.~Qu, X.~Yuan, M.-A. Côté, Z.~Liu, , and J.~Tang, ``Graph
  policy network for transferable active learning on graphs,'' in
  \emph{Advances in Neural Information Processing Systems (NIPS)}, 2020.

\bibitem{gammelli2021graph}
D.~Gammelli, K.~Yang, J.~Harrison, F.~Rodrigues, F.~C. Pereira, and M.~Pavone,
  ``Graph neural network reinforcement learning for autonomous
  mobility-on-demand systems,'' \emph{arXiv preprint arXiv:2104.11434}, 2021.

\bibitem{Wang2018NerveNet}
T.~Wang, R.~Liao, J.~Ba, and S.~Fidler, ``Nervenet: Learning structured policy
  with graph neural networks,'' in \emph{International Conference on Learning
  Representations (ICLR)}, 2018.

\bibitem{Johnson_2015_CVPR}
J.~Johnson, R.~Krishna, M.~Stark, L.-J. Li, D.~Shamma, M.~Bernstein, and
  L.~Fei-Fei, ``Image retrieval using scene graphs,'' in \emph{Proceedings of
  the IEEE Conference on Computer Vision and Pattern Recognition (CVPR)}, June
  2015.

\bibitem{Schroeder_2020_CVPR_Workshops}
B.~Schroeder and S.~Tripathi, ``Structured query-based image retrieval using
  scene graphs,'' in \emph{Proceedings of the IEEE/CVF Conference on Computer
  Vision and Pattern Recognition (CVPR) Workshops}, June 2020.

\bibitem{Wang_2020_WACV}
S.~Wang, R.~Wang, Z.~Yao, S.~Shan, and X.~Chen, ``Cross-modal scene graph
  matching for relationship-aware image-text retrieval,'' in \emph{Proceedings
  of the IEEE/CVF Winter Conference on Applications of Computer Vision (WACV)},
  March 2020.

\bibitem{johnson2018image}
J.~Johnson, A.~Gupta, and L.~Fei-Fei, ``Image generation from scene graphs,''
  in \emph{CVPR}, 2018.

\bibitem{herzigeccv2020}
R.~Herzig, A.~Bar, H.~Xu, G.~Chechik, T.~Darrell, and A.~Globerson, ``Learning
  canonical representations for scene graph to image generation,'' in
  \emph{European Conference on Computer Vision}, 2020.

\bibitem{Tripathi2019UsingSG}
S.~Tripathi, A.~Bhiwandiwalla, A.~Bastidas, and H.~Tang, ``Using scene graph
  context to improve image generation,'' \emph{ArXiv}, vol. abs/1901.03762,
  2019.

\bibitem{Ghosh2019GeneratingNL}
S.~Ghosh, G.~Burachas, A.~Ray, and A.~Ziskind, ``Generating natural language
  explanations for visual question answering using scene graphs and visual
  attention,'' \emph{ArXiv}, vol. abs/1902.05715, 2019.

\bibitem{Shi_2019_CVPR}
J.~Shi, H.~Zhang, and J.~Li, ``Explainable and explicit visual reasoning over
  scene graphs,'' in \emph{Proceedings of the IEEE/CVF Conference on Computer
  Vision and Pattern Recognition (CVPR)}, June 2019.

\bibitem{xu2017scenegraph}
D.~Xu, Y.~Zhu, C.~Choy, and L.~Fei-Fei, ``Scene graph generation by iterative
  message passing,'' in \emph{Computer Vision and Pattern Recognition (CVPR)},
  2017.

\bibitem{yang2018graph}
J.~Yang, J.~Lu, S.~Lee, D.~Batra, and D.~Parikh, ``Graph r-cnn for scene graph
  generation,'' in \emph{ECCV}, 2018.

\bibitem{Dai2017relation}
B.~Dai, Y.~Zhang, and D.~Lin, ``Detecting visual relationships with deep
  relational networks,'' in \emph{CVPR}, 2017.

\bibitem{dai2017detecting}
------, ``Detecting visual relationships with deep relational networks,'' in
  \emph{Proceedings of the IEEE Conference on Computer Vision and Pattern
  Recognition}, 2017.

\bibitem{Li_2017_ICCV}
Y.~Li, W.~Ouyang, B.~Zhou, K.~Wang, and X.~Wang, ``Scene graph generation from
  objects, phrases and region captions,'' in \emph{Proceedings of the IEEE
  International Conference on Computer Vision (ICCV)}, Oct 2017.

\bibitem{bresenham1965algorithm}
J.~E. Bresenham, ``Algorithm for computer control of a digital plotter,''
  \emph{IBM Systems journal}, vol.~4, no.~1, pp. 25--30, 1965.

\bibitem{kipf2016semi}
T.~N. Kipf and M.~Welling, ``Semi-supervised classification with graph
  convolutional networks,'' \emph{International Conference on Learning
  Representations (ICLR)}, 2017.

\bibitem{Velickovic18iclrGAT}
P.~Veli\v{c}kovi\'{c}, G.~Cucurull, A.~Casanova, A.~Romero, P.~Li\'{o}, and
  Y.~Bengio, ``Graph attention networks,'' in \emph{International Conference on
  Learning Representations (ICLR)}, May 2018.

\bibitem{hamilton2017GarphSage}
W.~L. Hamilton, R.~Ying, and J.~Leskovec, ``Inductive representation learning
  on large graphs,'' in \emph{NIPS}, 2017.

\bibitem{Konda00actorcriticalgorithms}
V.~Konda and J.~Tsitsiklis, ``Actor-critic algorithms,'' in \emph{SIAM Journal
  on Control and Optimization}.\hskip 1em plus 0.5em minus 0.4em\relax MIT
  Press, 2000, pp. 1008--1014.

\bibitem{schulman2017ppo}
J.~Schulman, F.~Wolski, P.~Dhariwal, A.~Radford, and O.~Klimov, ``Proximal
  policy optimization algorithms.'' \emph{CoRR}, vol. abs/1707.06347, 2017.

\bibitem{Mnih2015Nature}
V.~Mnih, K.~Kavukcuoglu, D.~Silver, A.~A. Rusu, J.~Veness, M.~G. Bellemare,
  A.~Graves, M.~A. Riedmiller, A.~Fidjeland, G.~Ostrovski, S.~Petersen,
  C.~Beattie, A.~Sadik, I.~Antonoglou, H.~King, D.~Kumaran, D.~Wierstra,
  S.~Legg, and D.~Hassabis, ``Human-level control through deep reinforcement
  learning.'' \emph{Nature}, vol. 518, no. 7540, pp. 529--533, 2015.

\bibitem{Hughes2022Hydra}
N.~Hughes, Y.~Chang, and L.~Carlone, ``Hydra: A real-time spatial perception
  engine for 3d scene graph construction and optimization,'' \emph{ArXiv}, vol.
  abs/2201.13360, 2022.

\bibitem{pytorch_geometric}
M.~Fey and J.~E. Lenssen, ``Fast graph representation learning with {PyTorch
  Geometric},'' in \emph{ICLR Workshop on Representation Learning on Graphs and
  Manifolds}, 2019.

\bibitem{pytorch}
A.~Paszke, S.~Gross, F.~Massa, A.~Lerer, J.~Bradbury, G.~Chanan, T.~Killeen,
  Z.~Lin, N.~Gimelshein, L.~Antiga, A.~Desmaison, A.~Kopf, E.~Yang, Z.~DeVito,
  M.~Raison, A.~Tejani, S.~Chilamkurthy, B.~Steiner, L.~Fang, J.~Bai, and
  S.~Chintala, ``Pytorch: An imperative style, high-performance deep learning
  library,'' in \emph{Advances in Neural Information Processing Systems 32},
  2019.

\bibitem{rllib}
\BIBentryALTinterwordspacing
E.~Liang, R.~Liaw, R.~Nishihara, P.~Moritz, R.~Fox, K.~Goldberg, J.~Gonzalez,
  M.~Jordan, and I.~Stoica, ``{RL}lib: Abstractions for distributed
  reinforcement learning,'' in \emph{Proceedings of the 35th International
  Conference on Machine Learning}, ser. Proceedings of Machine Learning
  Research, J.~Dy and A.~Krause, Eds., vol.~80.\hskip 1em plus 0.5em minus
  0.4em\relax PMLR, 10--15 Jul 2018, pp. 3053--3062. [Online]. Available:
  \url{http://proceedings.mlr.press/v80/liang18b.html}
\BIBentrySTDinterwordspacing

\bibitem{openai_gym}
G.~Brockman, V.~Cheung, L.~Pettersson, J.~Schneider, J.~Schulman, J.~Tang, and
  W.~Zaremba, ``Openai gym,'' 2016.

\bibitem{kipf2017iclr}
T.~N. Kipf and M.~Welling, ``Semi-supervised classification with graph
  convolutional networks,'' in \emph{International Conference on Learning
  Representations (ICLR)}, 2017.

\bibitem{bootstrap}
B.~Efron and R.~Tibshirani, \emph{An Introduction to the Bootstrap}.\hskip 1em
  plus 0.5em minus 0.4em\relax {Chapman \& Hall}, 1993.

\end{thebibliography}

\newpage
\appendices
\setcounter{figure}{0}
\setcounter{table}{0}
\renewcommand{\thefigure}{A\arabic{figure}}
\renewcommand{\thetable}{A\arabic{table}}
\renewcommand{\thesection}{A\arabic{section}}


\section{Summary}
\label{sec:appendix_summary}
This appendix provides additional implementation details and experimental results.
Section~\ref{sec:policy} provides details about the software libraries used,
2D ESDF (Euclidean Signed Distance Function) slice construction, baseline architectures, and hyperparameters.
Both an open-source implementation of the proposed method and the simulator used in our experiments have been released.~\footnote{\url{www.github.com/MIT-TESSE/dsg-rl}}
Section~\ref{sec:appendix_setup} discusses our experimental setup and provides visualizations of our simulation environments and corresponding Dynamic Scene Graphs (DSGs).
Section~\ref{sec:results} provides visualizations of the results from Section 4.2 (Comparison with Baselines) and
Section 4.3 (On the Importance of Hierarchy and Explicit Memory) in the main paper.
\highlight{Section~\ref{sec:sim2real} considers the effect of noise in the DSG creation on our method.
	Specifically, we run ablation studies that perturb node locations, inject semantic noise, and add latency to room node observations.}


\section{Implementation Details} 
\label{sec:policy}

\textbf{Software Libraries Used.}
We use PyTorch Geometric~\cite{pytorch_geometric} to implement our Graph Neural Networks (GNNs).
Other learned components (\eg Convolutional Neural Networks (CNNs), Multi-layer Perceptrons (MLPs)) are built with PyTorch~\cite{pytorch}.
We use RLlib's~\cite{rllib} implementation of Proximal Policy Optimization~\cite{schulman2017ppo} for policy training and evaluation.
We implement our object search task using the OpenAI Gym~\cite{openai_gym} framework.

\textbf{Policy Architecture.}
We experiment with various GNN iterations $K$ (\eg 2, 3, 4) and hidden dimensions (\eg 32, 64, 128) and find 3 iterations of 64 hidden dimensions to be optimal.
All layers of our GNN use the Graph Convolutional Operator~\cite{kipf2017iclr}, and we defer an exploration of more complex architectures to future work.

\textbf{2D ESDF Slice.}
In order to efficiently perform occupancy checking as described in Section 3.2 of the main paper, we extract a 2D slice from the full ESDF (Euclidean Signed Distance Function).
To extract this slice, we first consider ESDF voxels with z coordinates corresponding the robots's vertical profile plus a small buffer;
this vertical profile is 0.475\,m to 0.925\,m in our experiments.
Each pixel in the 2D ESDF thus corresponds to a set of voxels of varying heights and the same x and y coordinates as that pixel.
We then populate each pixel with the minimum value across the corresponding voxels.
The resulting 2D ESDF slice provides occupancy information useful for robot navigation in the 2D plane (recall that our action space consists of ``move forward'', ``turn left'', ``turn right'' actions).

\textbf{Baseline Architectures.}
Similarly to our policy architecture discussed in Section 3.3 of the main paper, our baseline policies use an Actor-Critic method with an identical backbone function $f(o^t)$ where $o^t$ is the observation at time $t$ as discussed in Section 4.2 of the main paper.

Backbone function $f(o^t)$ is comprised of a Convolutional Neural Network (CNN), Multi-Layer Perception (MLP), and Recurrent Neural Network (RNN).
The CNN is modeled after the AtariNet~\cite{Mnih2015Nature} architecture.
All input images have a height and width of 120 and 160 pixels, respectively.
The CNN has three convolutional layers with kernel sizes 8$\times$8, 4$\times$4, 3$\times$3, channels 32, 64, 64, strides 4, 2, 1, and padding 4, 1, 1. Each convolutional layer is followed by a ReLU activation function.
The CNN output is flattened and then passed to an MLP of length 512.
The output feature vector is concatenated with ground truth pose which consists of the robot's x position, y position, and heading, and is then given to a Gated Recurrent Unit with a hidden dimension of 256.
The resulting feature vector is used by the actor and critic networks, each consisting of a single-layer MLP mapping the feature vector to an action distribution and return estimate, respectively.

\textbf{Hyperparameters.}
We start with RLlib's default PPO-specific configuration,\footnote{\url{https://docs.ray.io/en/releases-1.2.0/rllib-algorithms.html?highlight=ppo\#proximal-policy-optimization-ppo}}
then focus our hyperparameter tuning on the following parameters: learning rate, batch size, SGD (stochastic gradient descent) minibatch size, number of SGD iterations, discount factor, and GAE (Generalized Advantage Estimator) parameter.
We find that the same discount factor and GAE parameter of 0.99 and 0.9 is optimal across models.
We list the remaining parameters in Table~\ref{tab:hp}; we generally find that similar parameter ranges work well across approaches. Notably, a relatively large batch size is important for stable GNN training.

\begin{table*}[h]
	\centering
	\begin{tabular}{c c c c c c }
		\toprule
		Method            & Learning Rate & Batch Size & SGD Minibatch Size & Num. SGD Iter. \\
		\midrule
		Proposed Method   & 1e-4          & 4096       & 128                & 30             \\
		RGB-D + Semantics & 1e-5          & 512        & 128                & 8              \\
		RGB               & 1e-5          & 512        & 128                & 8              \\
		ESDF              & 1e-4          & 512        & 128                & 8              \\
		\bottomrule
	\end{tabular}
	\caption{Relevant hyperparameters used in the proposed approach and compared baselines.}
	\label{tab:hp}
\end{table*}


\newcommand{\envvspace}{2cm}
\newcommand{\fpvspace}{0.4}  

\section{Experimental Setup} 
\label{sec:appendix_setup}

\textbf{Evaluation Procedure.}
We run evaluation as detailed in Section 4.1 of the main paper.
During evaluation, we fix random seeds to keep relevant variables (\ie initial robot pose, target locations) constant across episodes.
Confidence intervals are estimated via the Bootstrap Method~\cite{bootstrap}, a resampling method that provides upper and lower performance bounds with 95\% confidence.
Note that we learn stochastic policies, thus during evaluation actions are sampled from the predicted action distribution.

\highlight{
	\textbf{Robot Description.}
	The robot in our setup is equipped with sensors that provide ground truth RGB imagery, 2D semantic segmentation, depth, and inertial data.
	The robot can execute one of three actions at each step: move forward 0.5\,m, turn left 8 degrees, or turn right 8 degrees.
	Finally, we model collisions with the environment using a capsule collider.
	The capsule has a height of 0.5\,m and a radius of 0.125\,m, and hovers above the ground with a 0.25\,m ground clearance.
}

\begin{figure}[h]
	\begin{minipage}{\textwidth}
		\centering
		{\includegraphics[width=\fpvspace\textwidth]{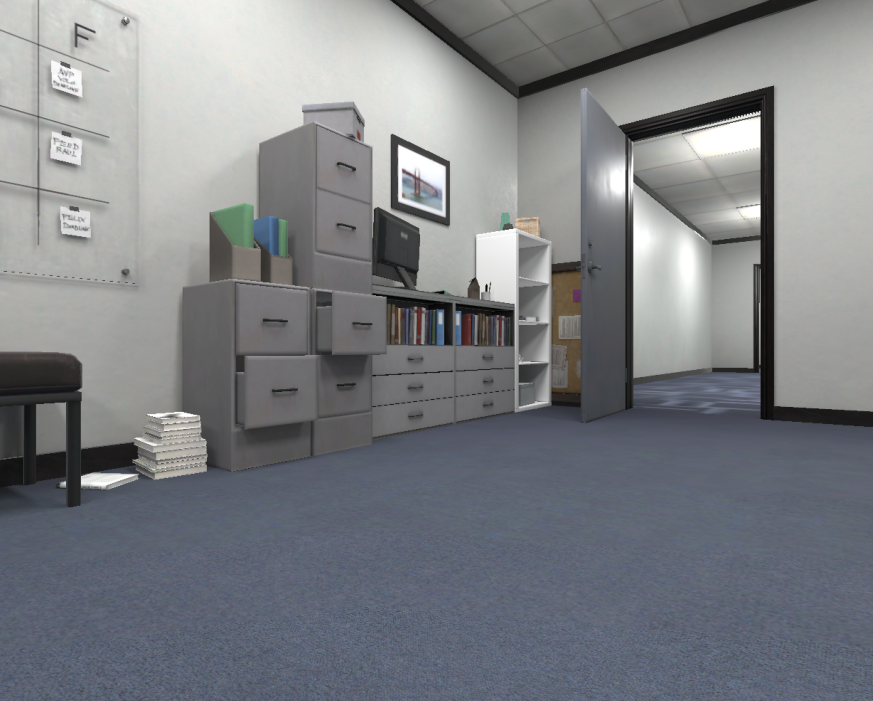}}
		\hfill
		{\includegraphics[width=\fpvspace\textwidth]{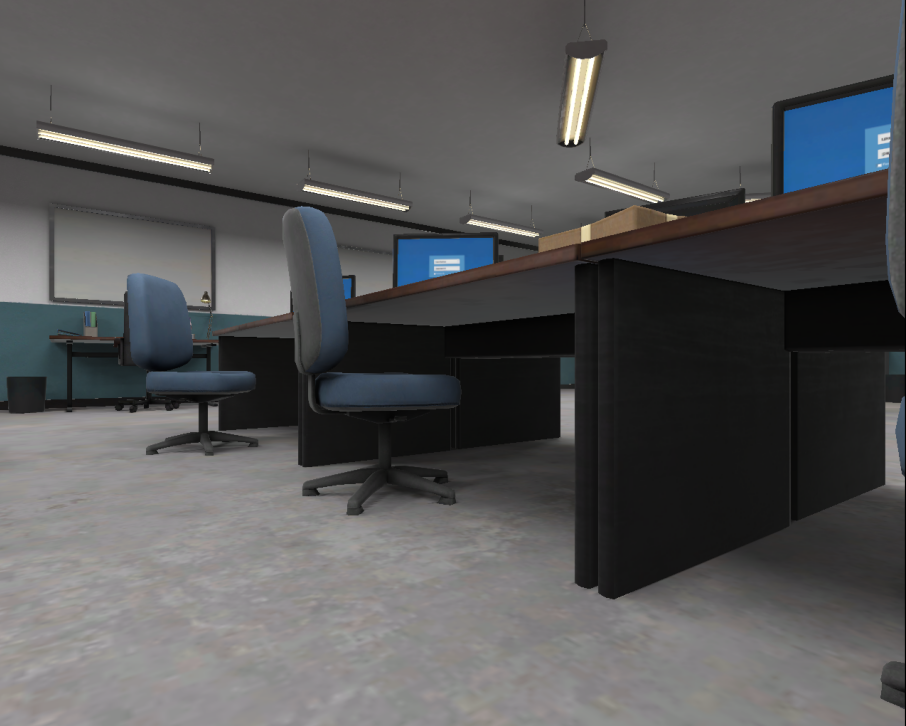}}
		\\
		{(a) \hspace{0.55\textwidth} (b)}
		\\
		\vspace{0.1in}
		{\includegraphics[width=\fpvspace\textwidth]{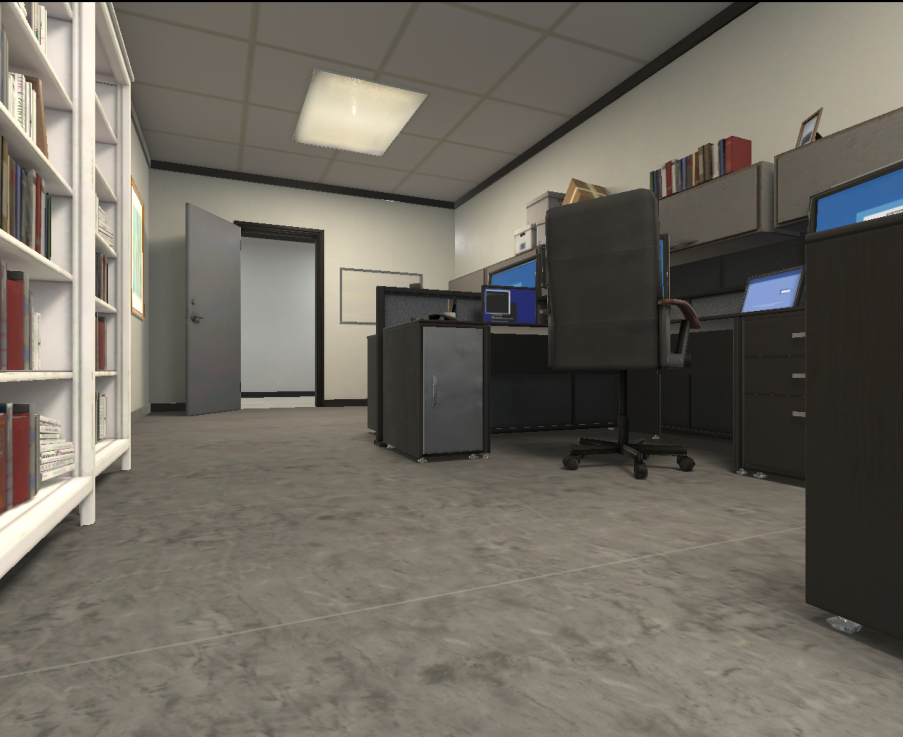}}
		\hfill
		{\includegraphics[width=\fpvspace\textwidth]{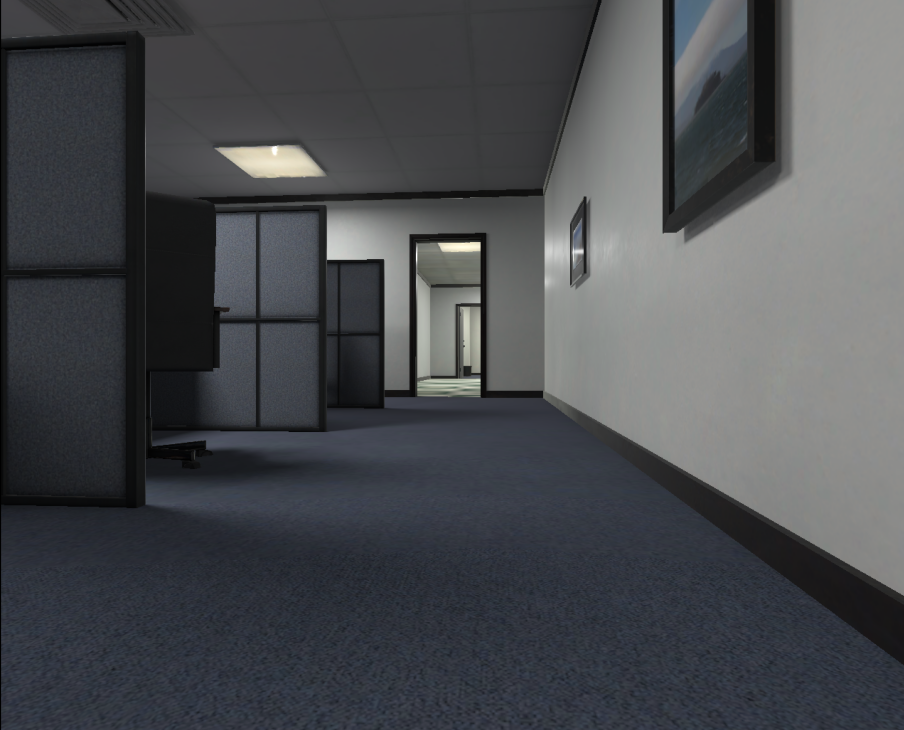}}
		\\
		{(c) \hspace{0.55\textwidth} (d)}
		\caption{Images taken from the robot's onboard RGB camera.}
		\label{fig:fpv}
	\end{minipage}
\end{figure}

\begin{figure*}[!ht]
	\centering
	\begin{minipage}{1.8\columnwidth}
		\includegraphics[width=\textwidth, trim=0 100 0 0, clip]{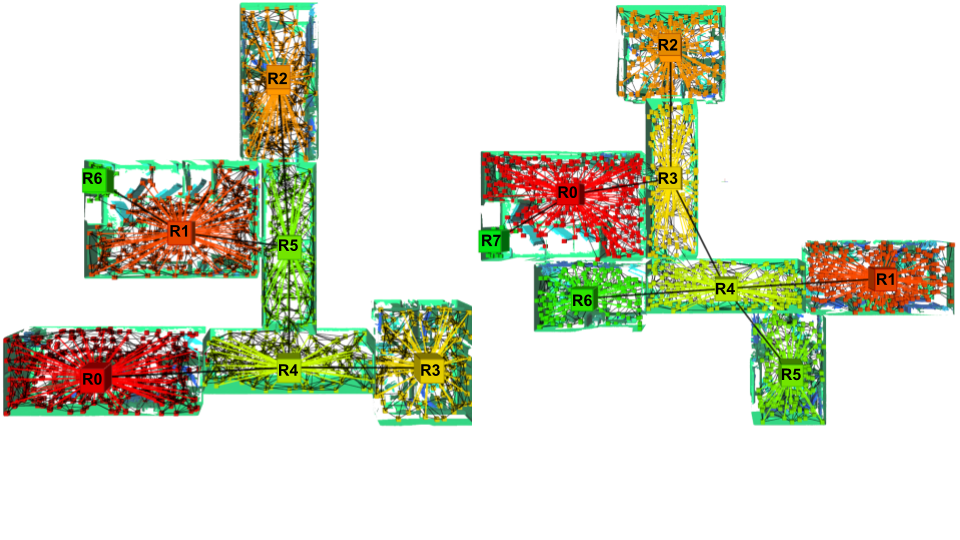} \\
		\hspace*{0.15\columnwidth}(a) DSG of Environment 1\hspace{0.25\columnwidth}(b) DSG of Environment 2
	\end{minipage}
	\\
	\vspace{\envvspace}
	\begin{minipage}{1.8\columnwidth}
		\includegraphics[width=\textwidth, trim=0 25 0 0, clip]{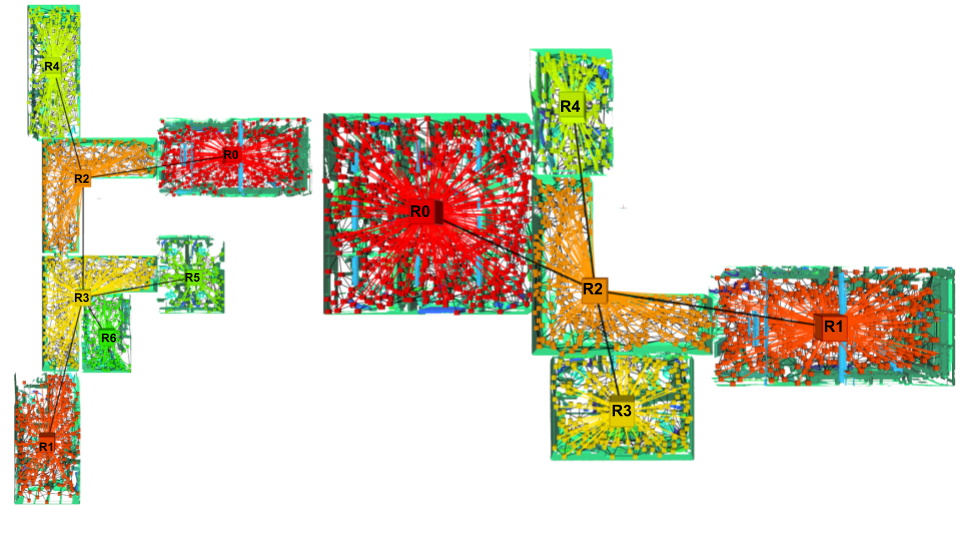} \\
		\hspace*{0.15\columnwidth}(c) DSG of Environment 3\hspace{0.25\columnwidth}(d) DSG of Environment 4
	\end{minipage}
	\caption{Top-down view of DSGs for Environments 1--4 showing the Room,\,Place,\,and\,Mesh\,layers. 
		\label{fig:dsg_topdown1}}
\end{figure*}

\begin{figure*}[!ht]
	\centering
	\begin{minipage}{1.8\columnwidth}
		\includegraphics[width=\textwidth, trim=0 80 0 0, clip]{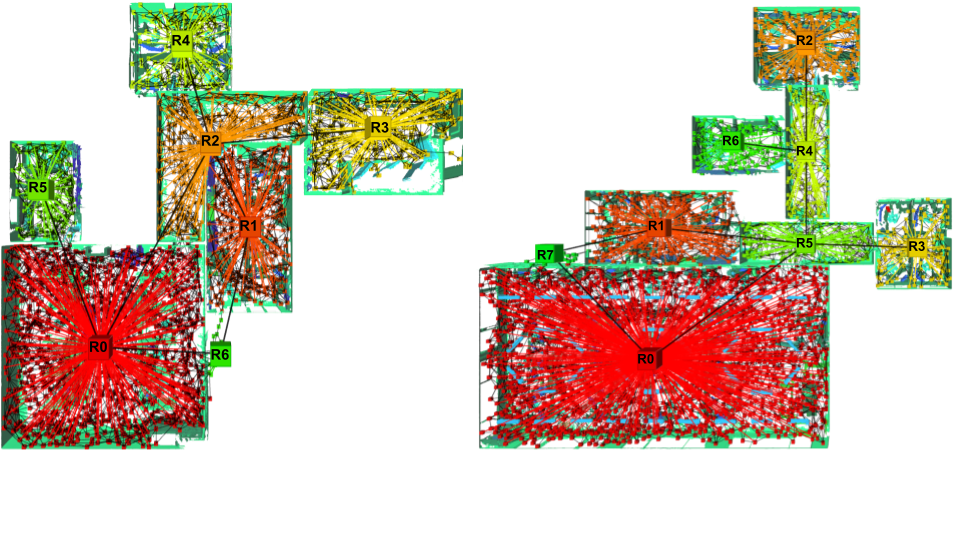} \\
		\hspace*{0.15\columnwidth}(a) DSG of Environment 5\hspace{0.25\columnwidth}(b) DSG of Environment 6
	\end{minipage}
	\\
	\vspace{\envvspace}
	\begin{minipage}{2\columnwidth}
		\centering
		\includegraphics[height=3in, trim=0 0 550 0, clip, scale=2]{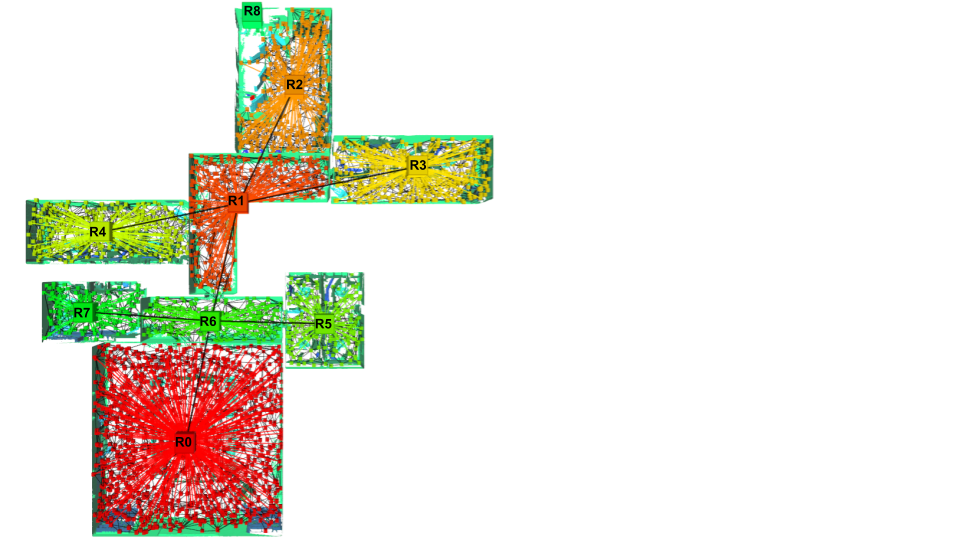} \\
		(c) DSG of Environment 7
	\end{minipage}
	\caption{Top-down view of DSGs for Environments 5--7 showing the Room,\,Place,\,and\,Mesh\,layers.
		\label{fig:dsg_topdown2}}
\end{figure*}

\textbf{Visualization of Environments.}
Figure~\ref{fig:fpv} provides images taken from the robot's onboard RGB camera.
Figure~\ref{fig:topdown1} and Figure~\ref{fig:topdown2} provide top-down visualizations of the environments, their areas in square meters, and the train-test split.
Figure~\ref{fig:dsg_topdown1} and Figure~\ref{fig:dsg_topdown2} show the corresponding DSGs.

\newcommand{\envwidth}{0.7}

\begin{figure*}[!ht]
	\centering
	\subfloat[Environment 1: 818\,m$^2$]
	{\includegraphics[width=\envwidth\textwidth]{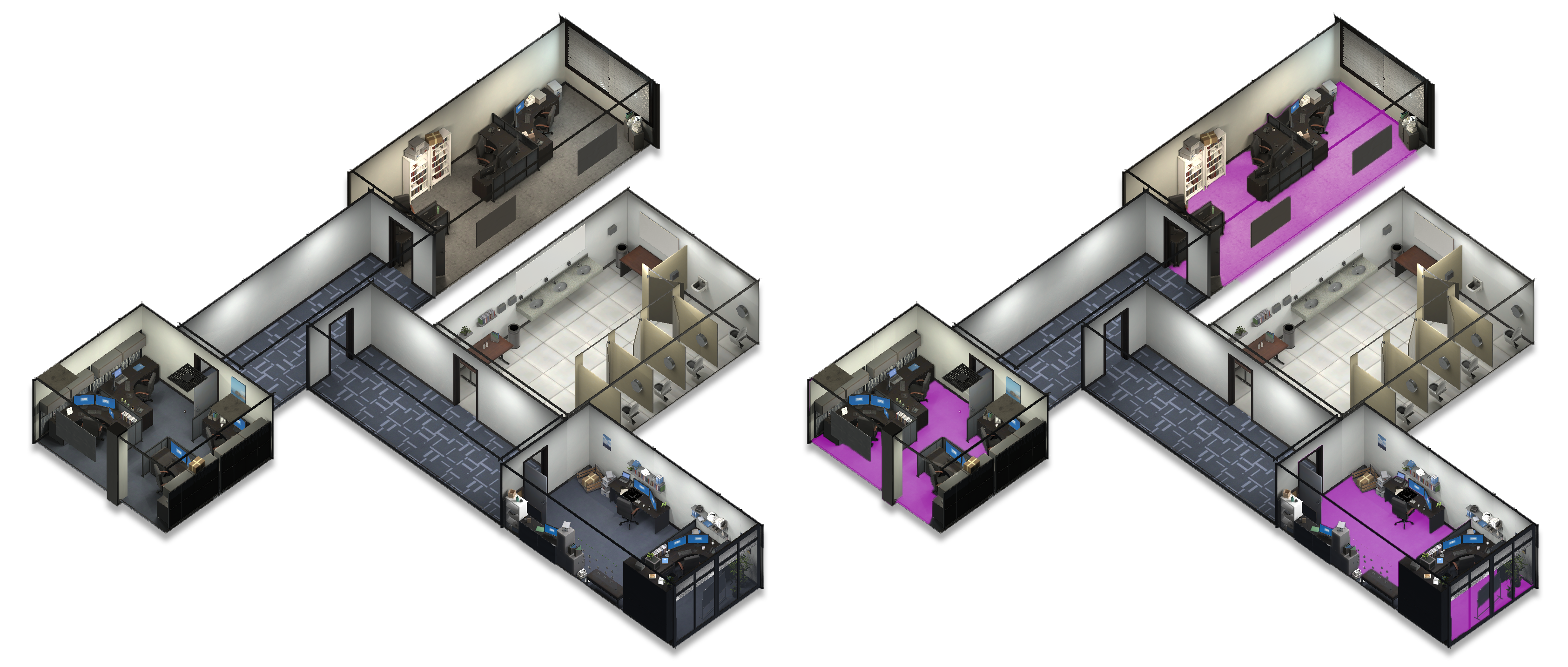}}
	\\
	\subfloat[Environment 2: 844\,m$^2$]
	{\includegraphics[width=\envwidth\textwidth]{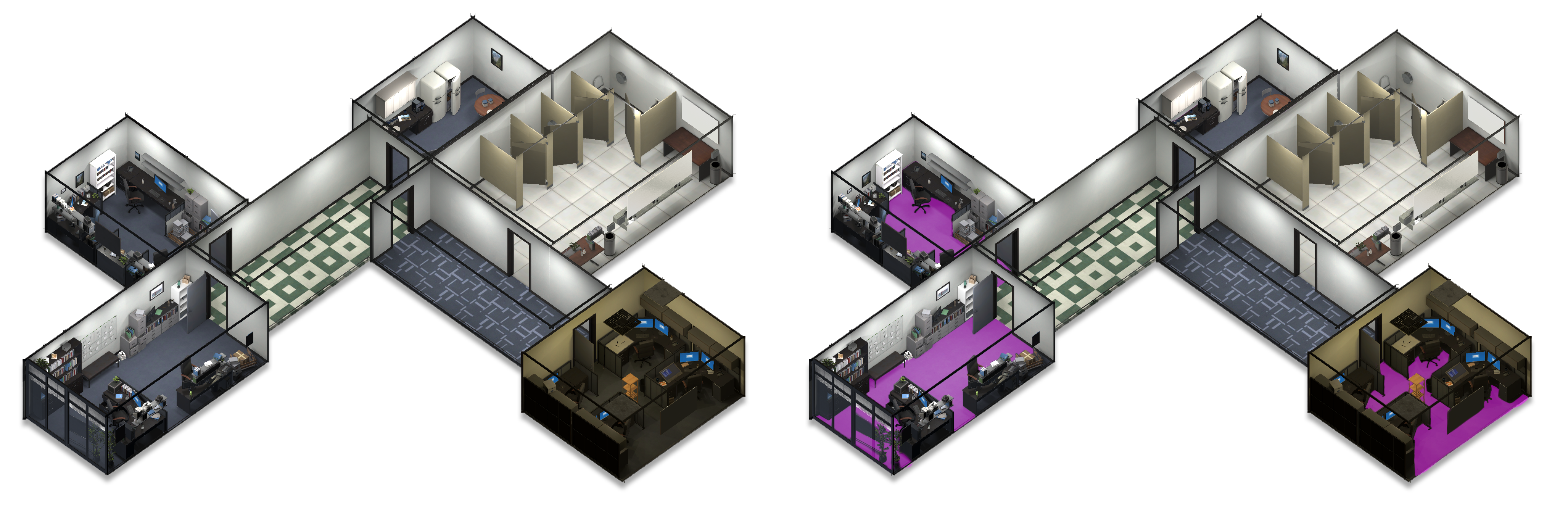}}
	\\
	\subfloat[Environment 3: 1167\,m$^2$]
	{\includegraphics[width=\envwidth\textwidth]{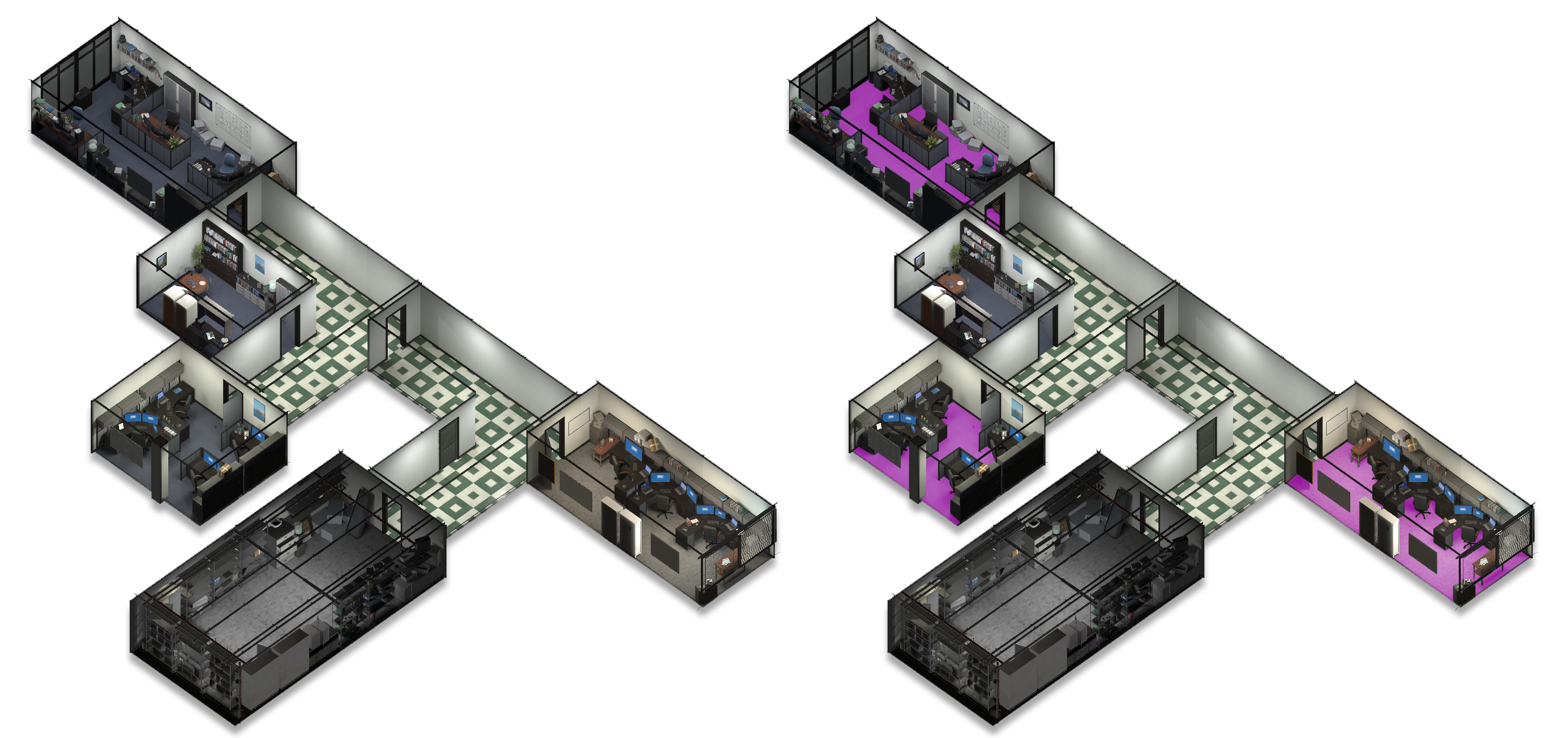}}
	\\
	\subfloat[Environment 4: 1059\,m$^2$]
	{\includegraphics[width=\envwidth\textwidth]{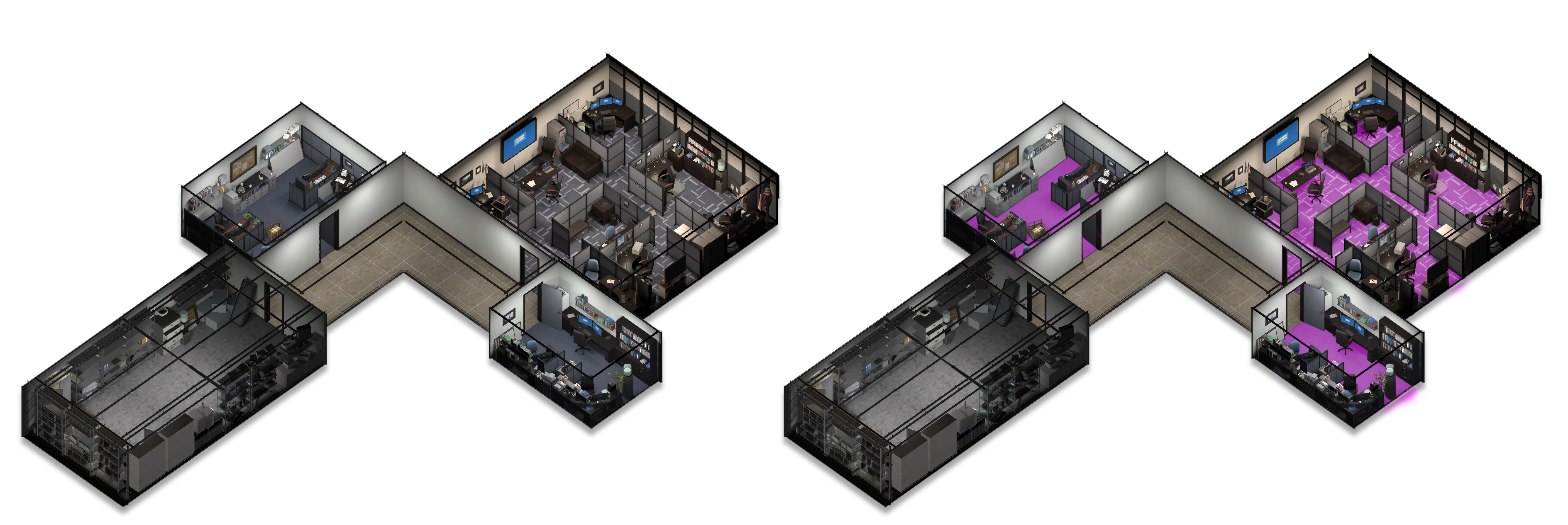}}
	\caption{Photo-realistic simulation environments 1-4, which are used for training.
		\highlight{Figures on the right-hand side highlight rooms where targets are randomly spawned.}}
	\label{fig:topdown1}
\end{figure*}

\begin{figure*}[!ht]
	\centering
	\subfloat[Environment 5: 1407\,m$^2$]
	{\includegraphics[width=\envwidth\textwidth]{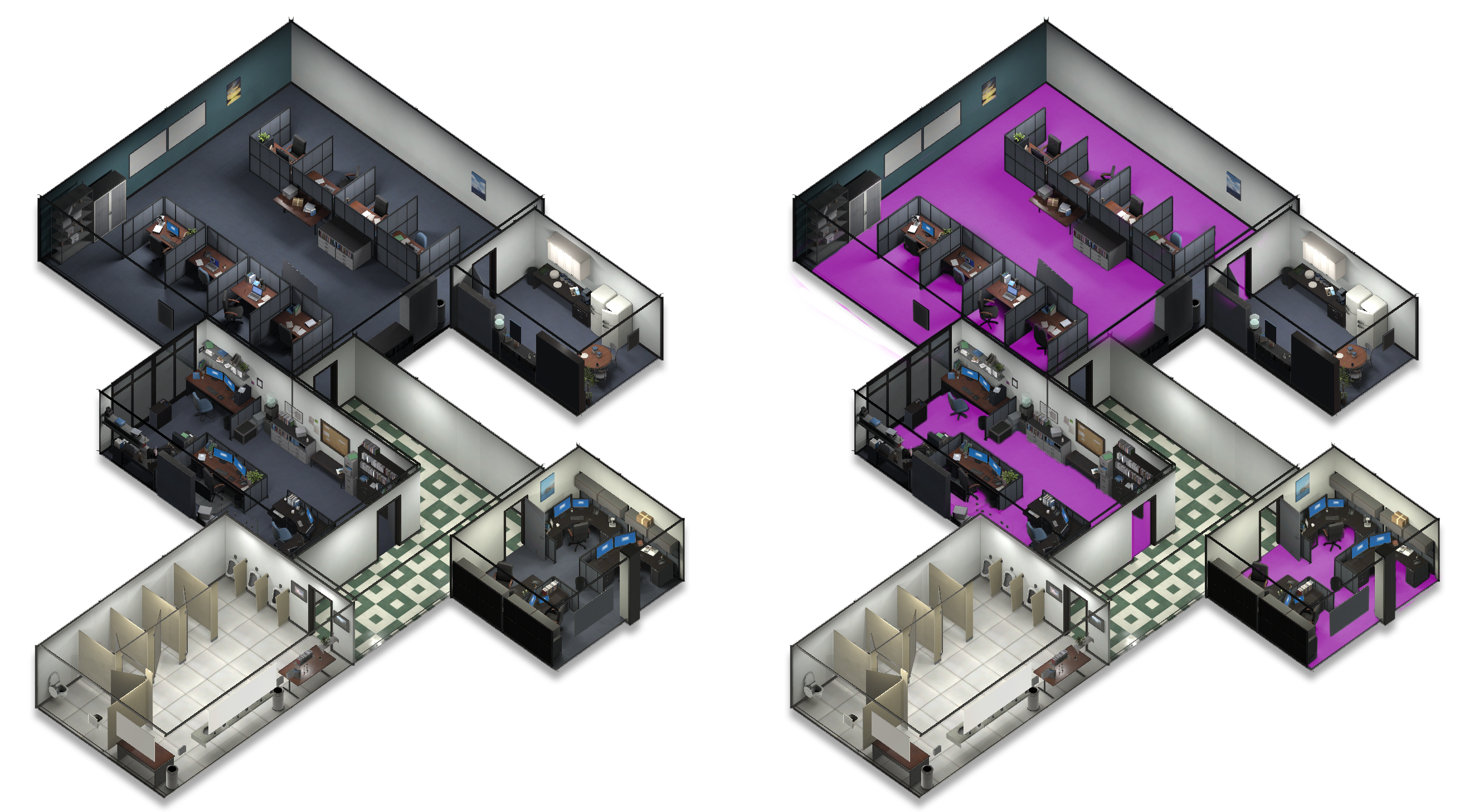}}
	\\
	\subfloat[Environment 6: 1882\,m$^2$]
	{\includegraphics[width=\envwidth\textwidth]{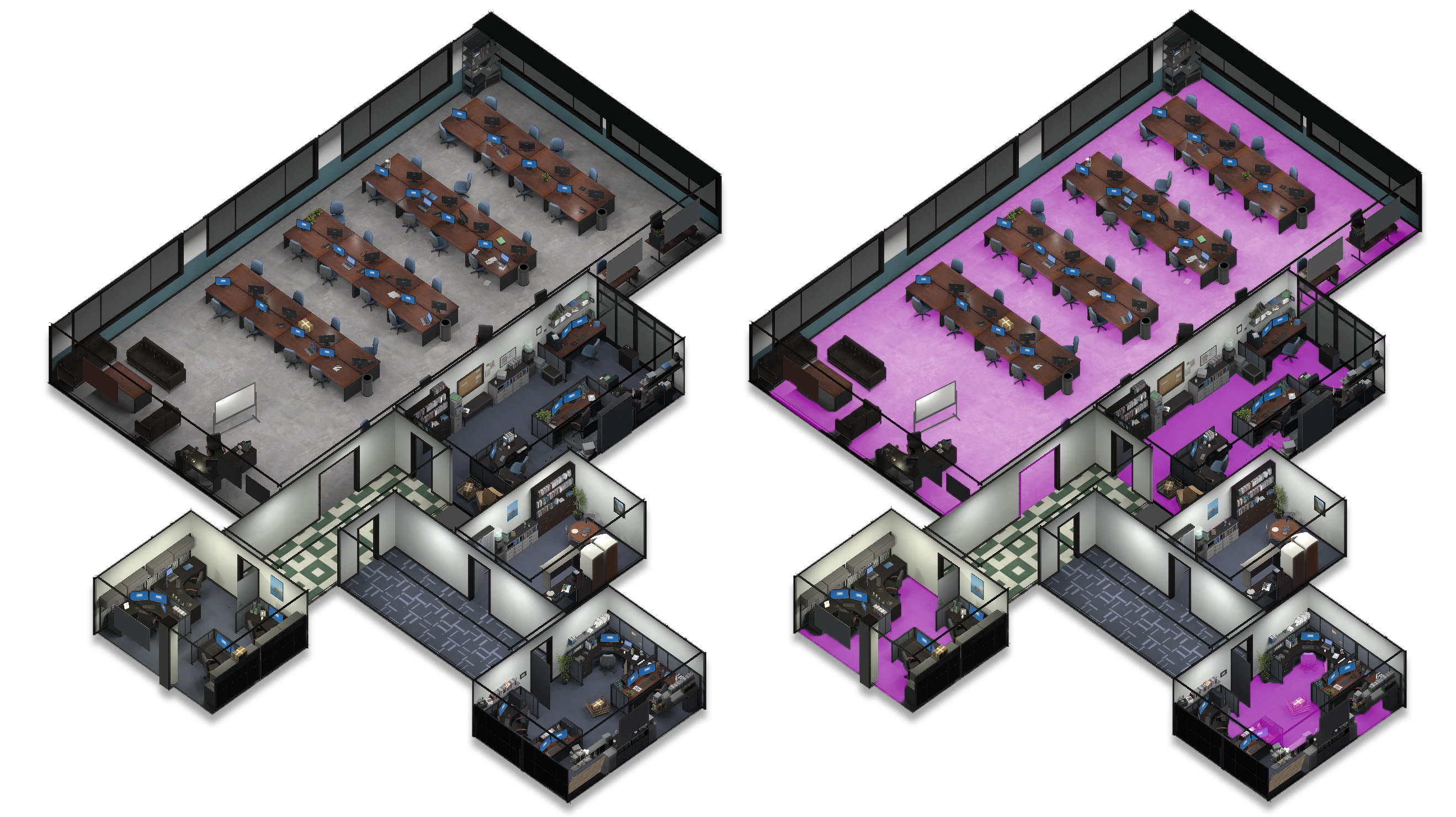}}
	\\
	\subfloat[Environment 7: 1567\,m$^2$]
	{\includegraphics[width=\envwidth\textwidth]{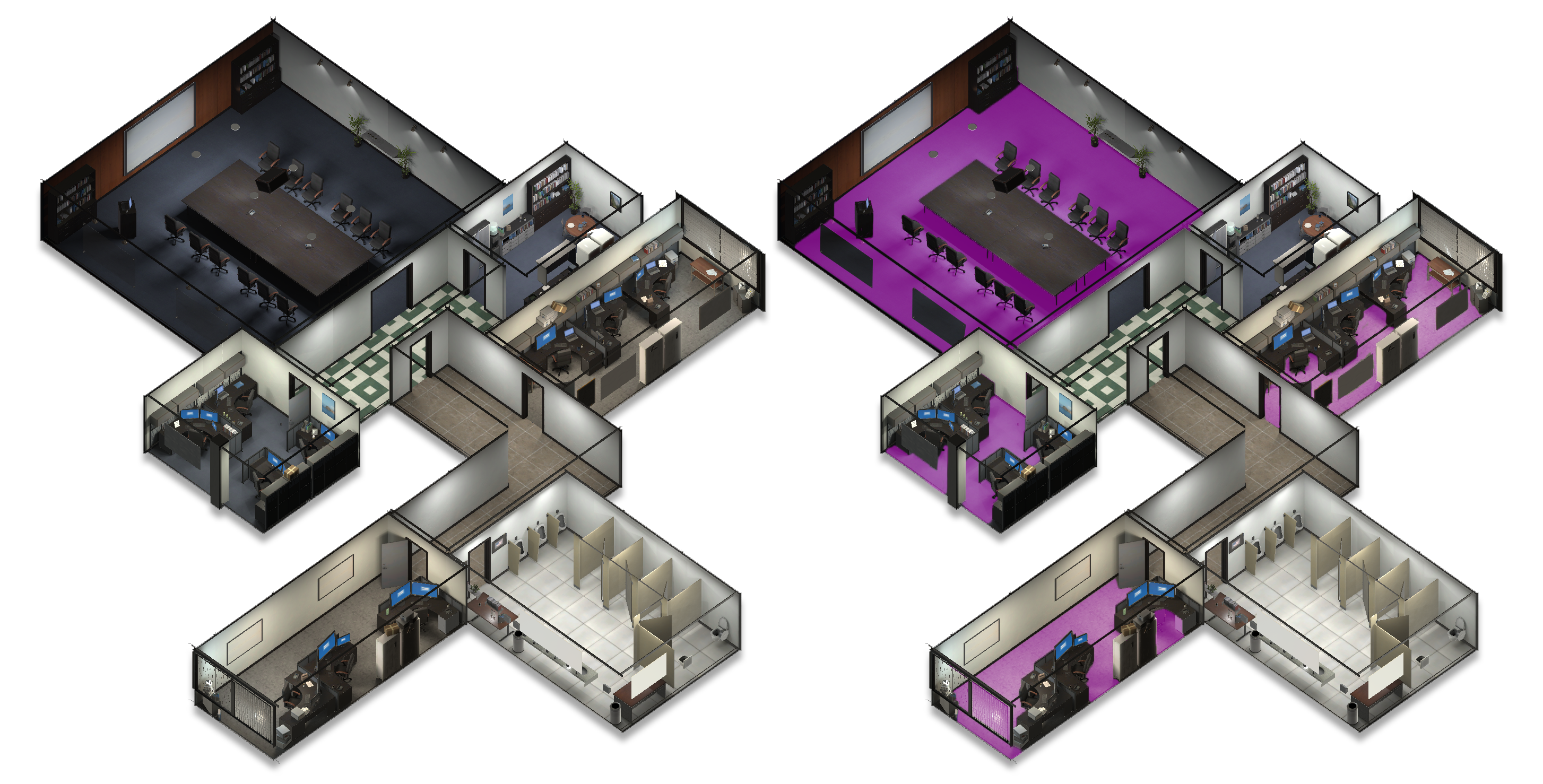}}
	\caption{Photo-realistic simulation environments 5-7, which are used for testing.
		\highlight{Figures on the right-hand side highlight rooms where targets are randomly spawned.}}
	\label{fig:topdown2}
\end{figure*}

\newcommand{\figwidth}{0.475}

\clearpage

\section{Additional Experimental Results}
\label{sec:results}
\begin{figure*}[]
	\centering
	\subfloat[]
	{\includegraphics[width=\figwidth\textwidth, trim=25 5 25 10]{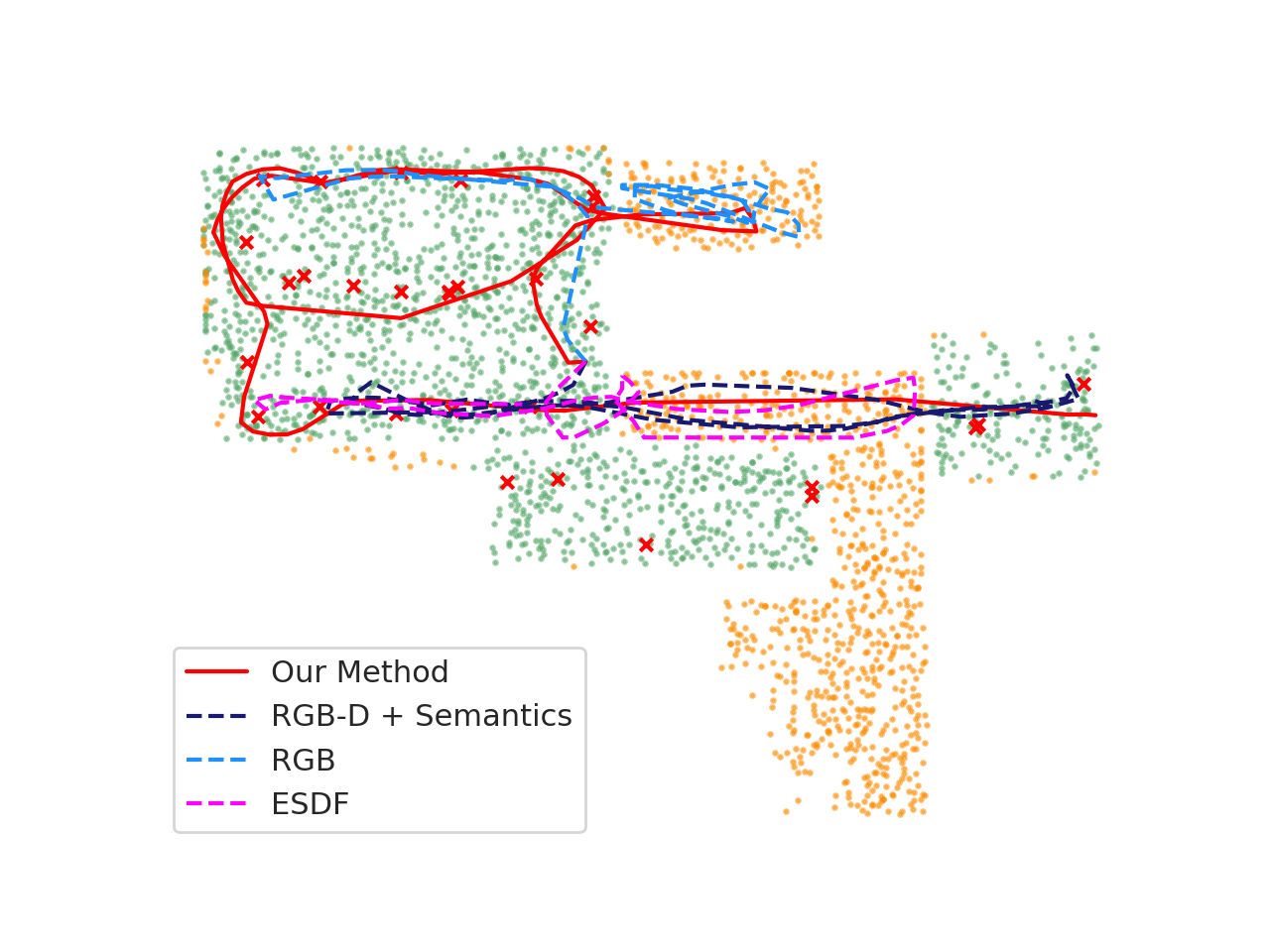}}
	\hfill
	\subfloat[]
	{\includegraphics[width=\figwidth\textwidth, trim=25 10 25 10]{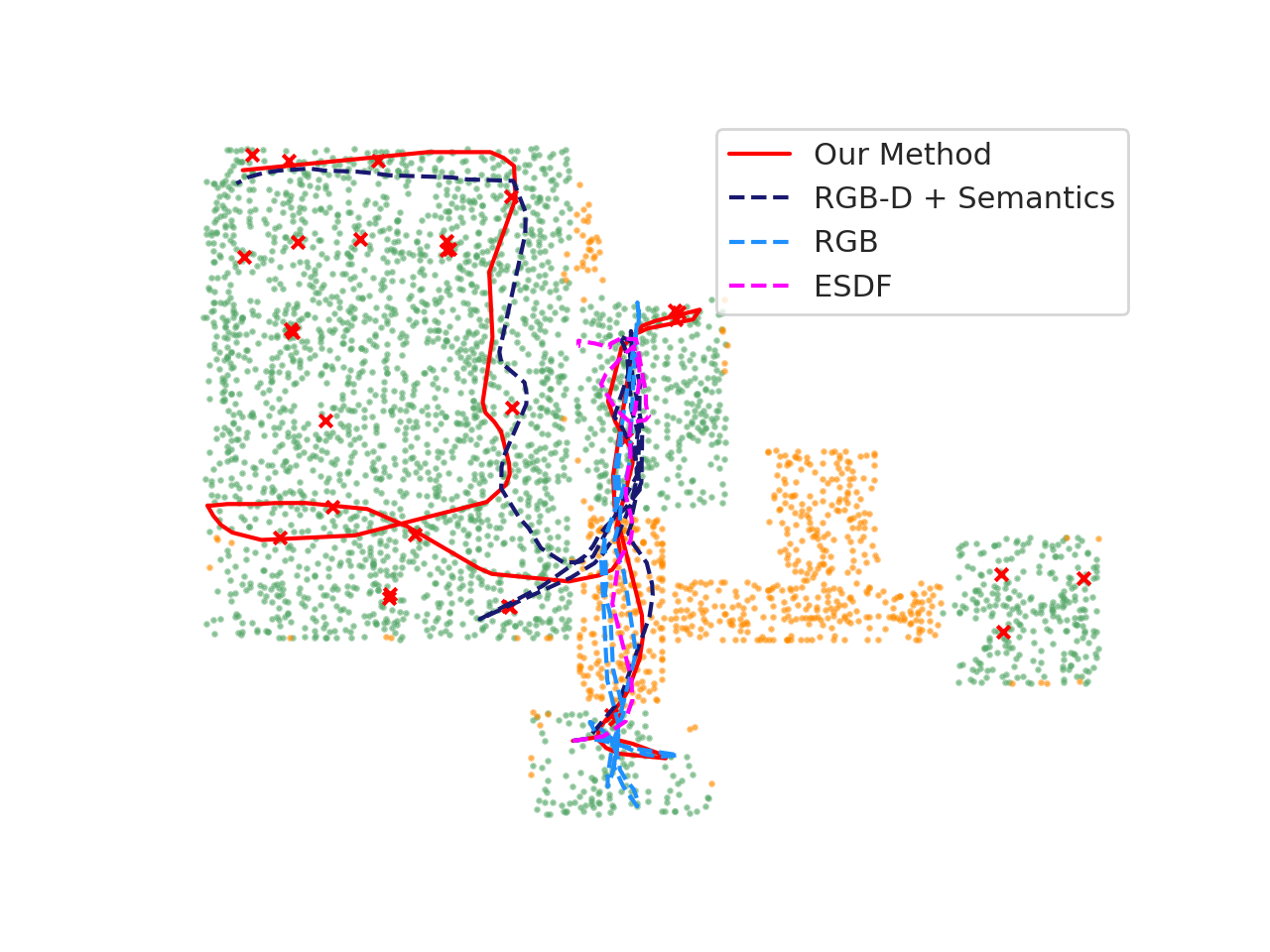}}
	\\
	\subfloat[]
	{\includegraphics[width=\figwidth\textwidth, trim=25 10 25 10]{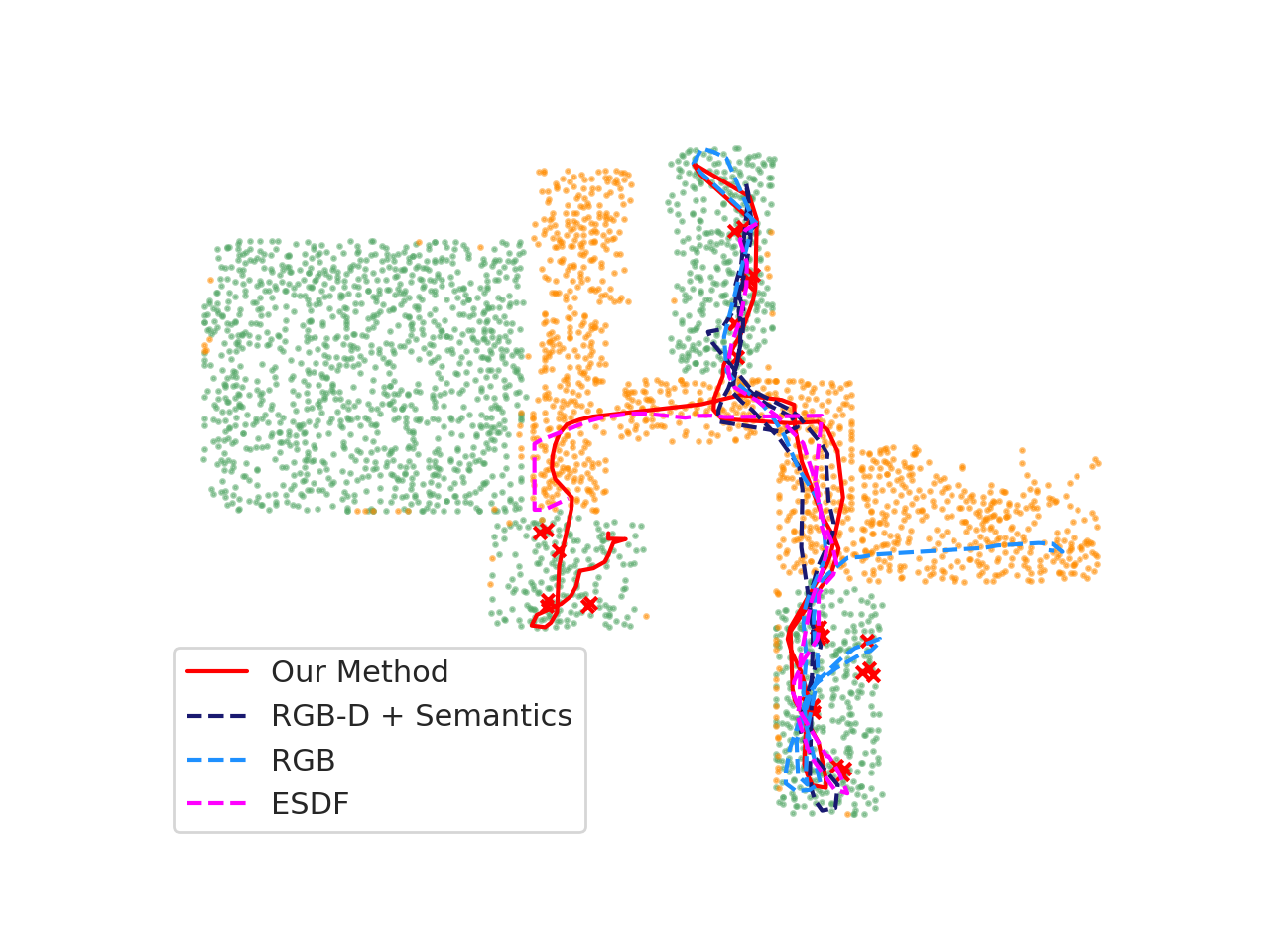}}
	\hfill
	\subfloat[]
	{\includegraphics[width=\figwidth\textwidth, trim=25 10 25 10]
		{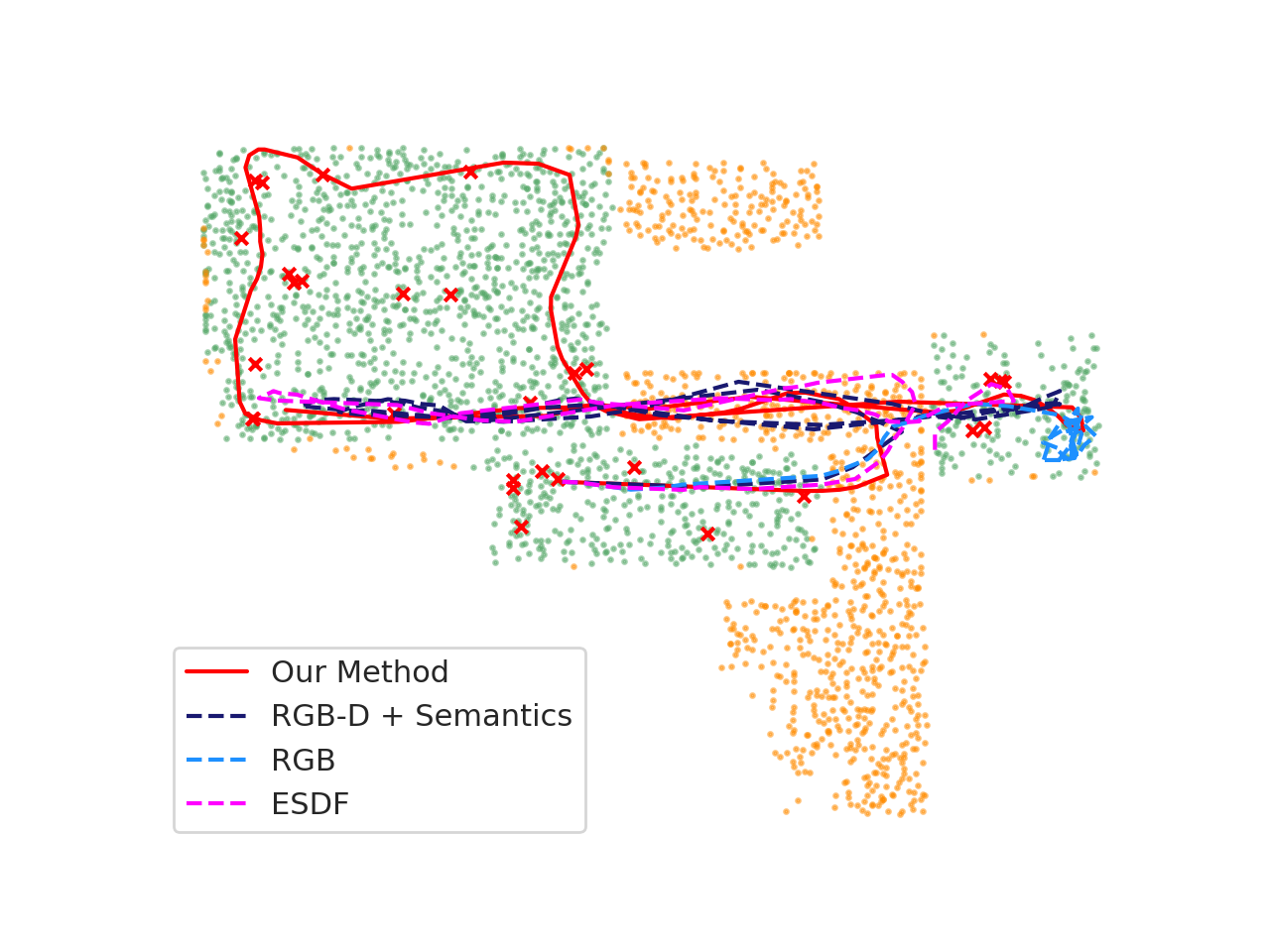}}
	\\
	\subfloat[]
	{\includegraphics[width=\figwidth\textwidth, trim=25 10 25 10]{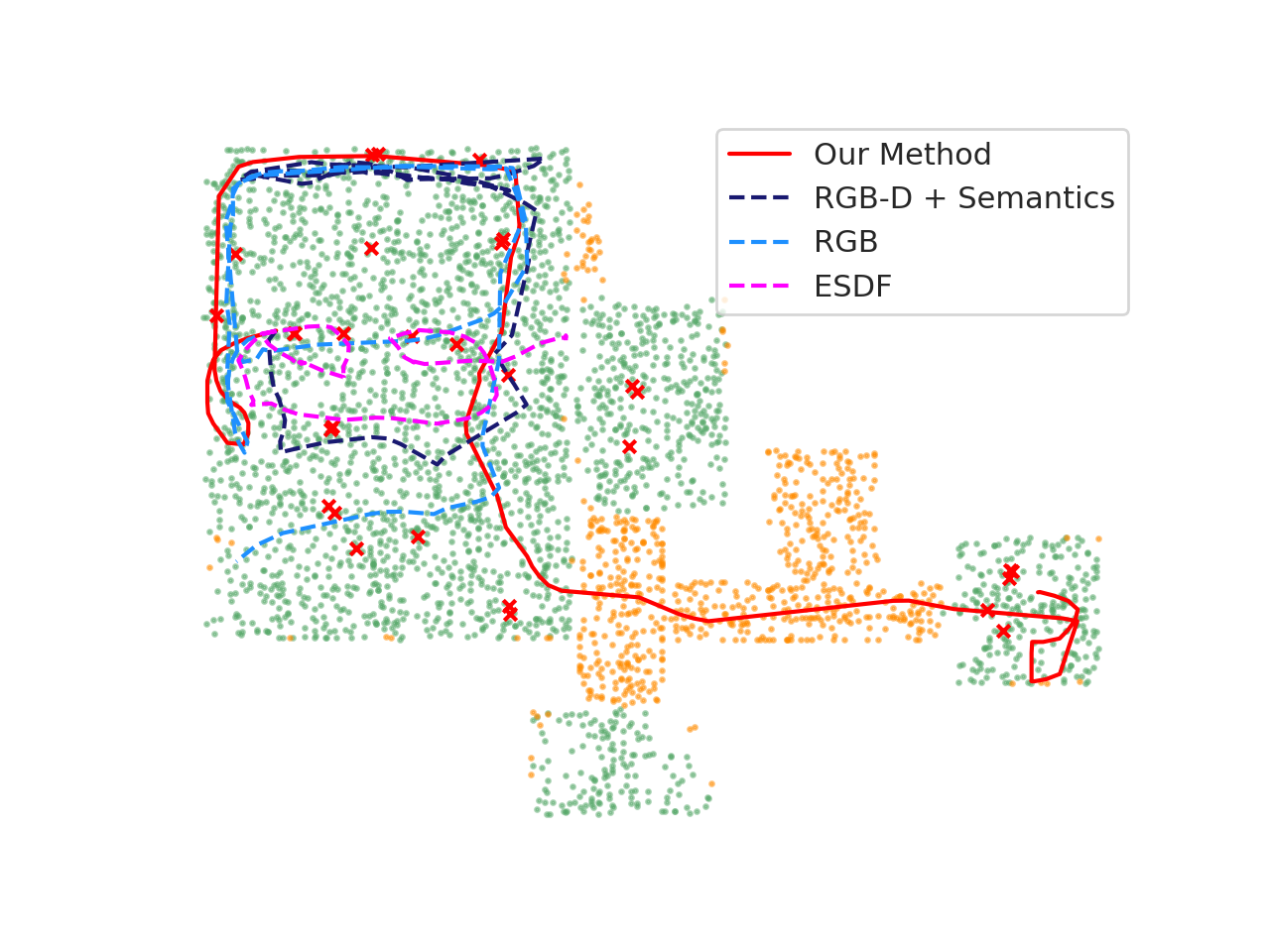}}
	\hfill
	\subfloat[]
	{\includegraphics[width=\figwidth\textwidth, trim=25 10 25 10]{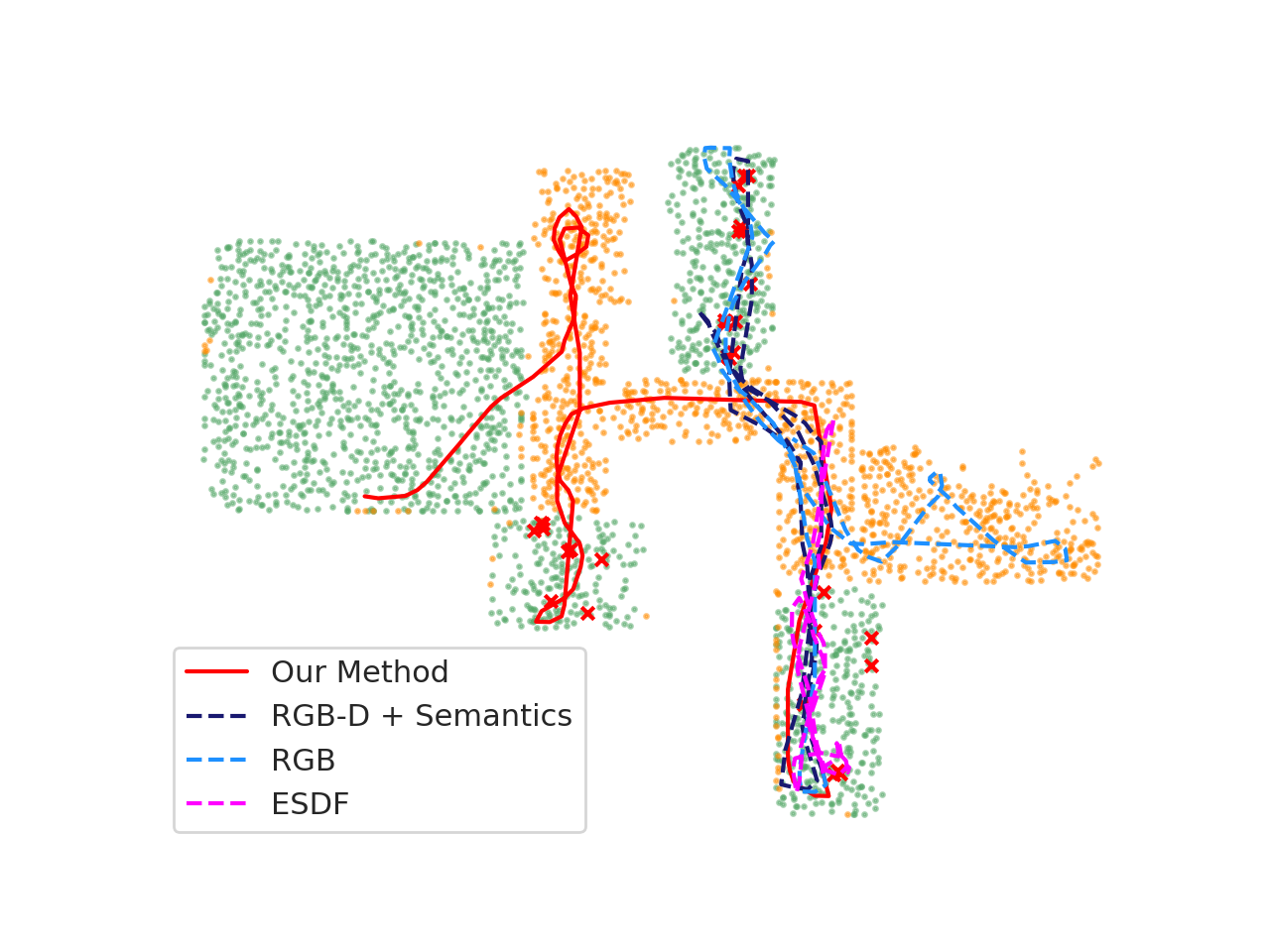}}
	\caption{Example trajectories comparing our method to policies using RGB, RGB-D + Semantics, and an ESDF slice, \highlight{with targets shown in red}. Rooms containing targets are labeled green, distractor rooms are orange. Our method spends more time exploring target rooms while minimizing revisited space.
		See section 4.1 of the main paper for quantitative results.}
	\label{fig:baseline-plots}
\end{figure*}

\begin{figure*}[]
	\centering
	\subfloat[]
	{\includegraphics[width=\figwidth\textwidth, trim=25 10 25 10]{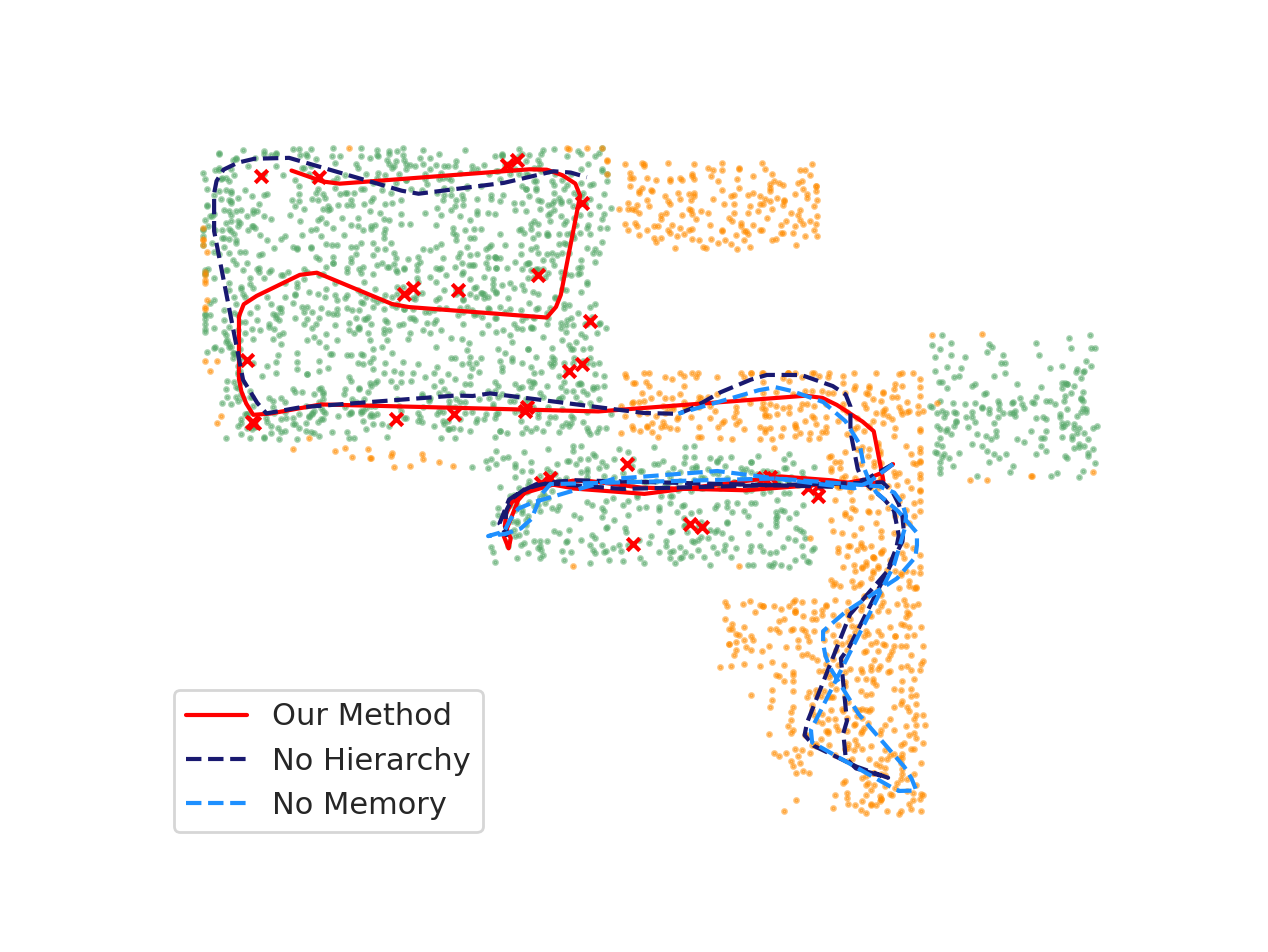}}
	\hfill
	\subfloat[]
	{\includegraphics[width=\figwidth\textwidth, trim=25 10 25 10]{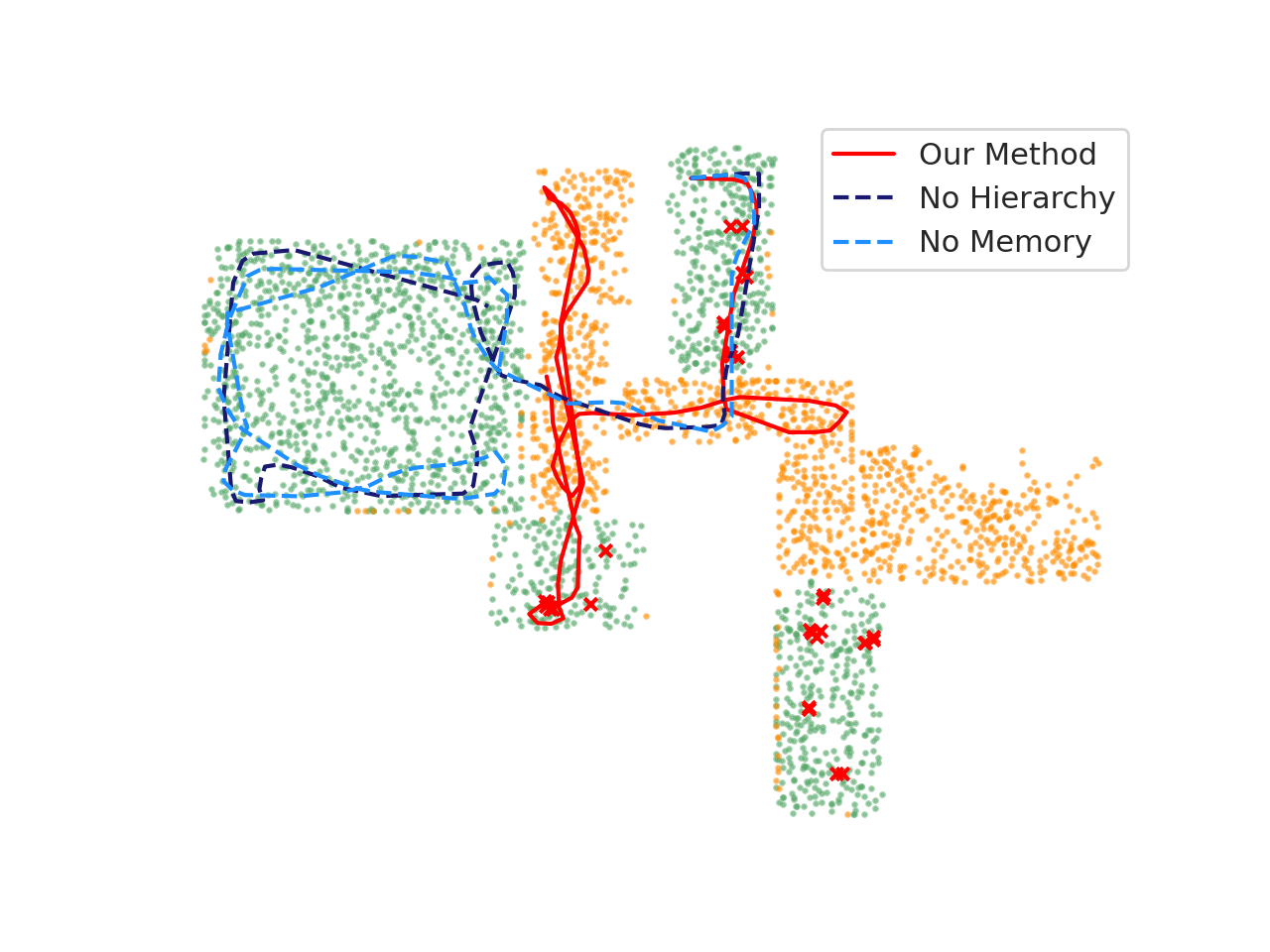}}
	\\
	\subfloat[]
	{\includegraphics[width=\figwidth\textwidth, trim=25 10 25 10]{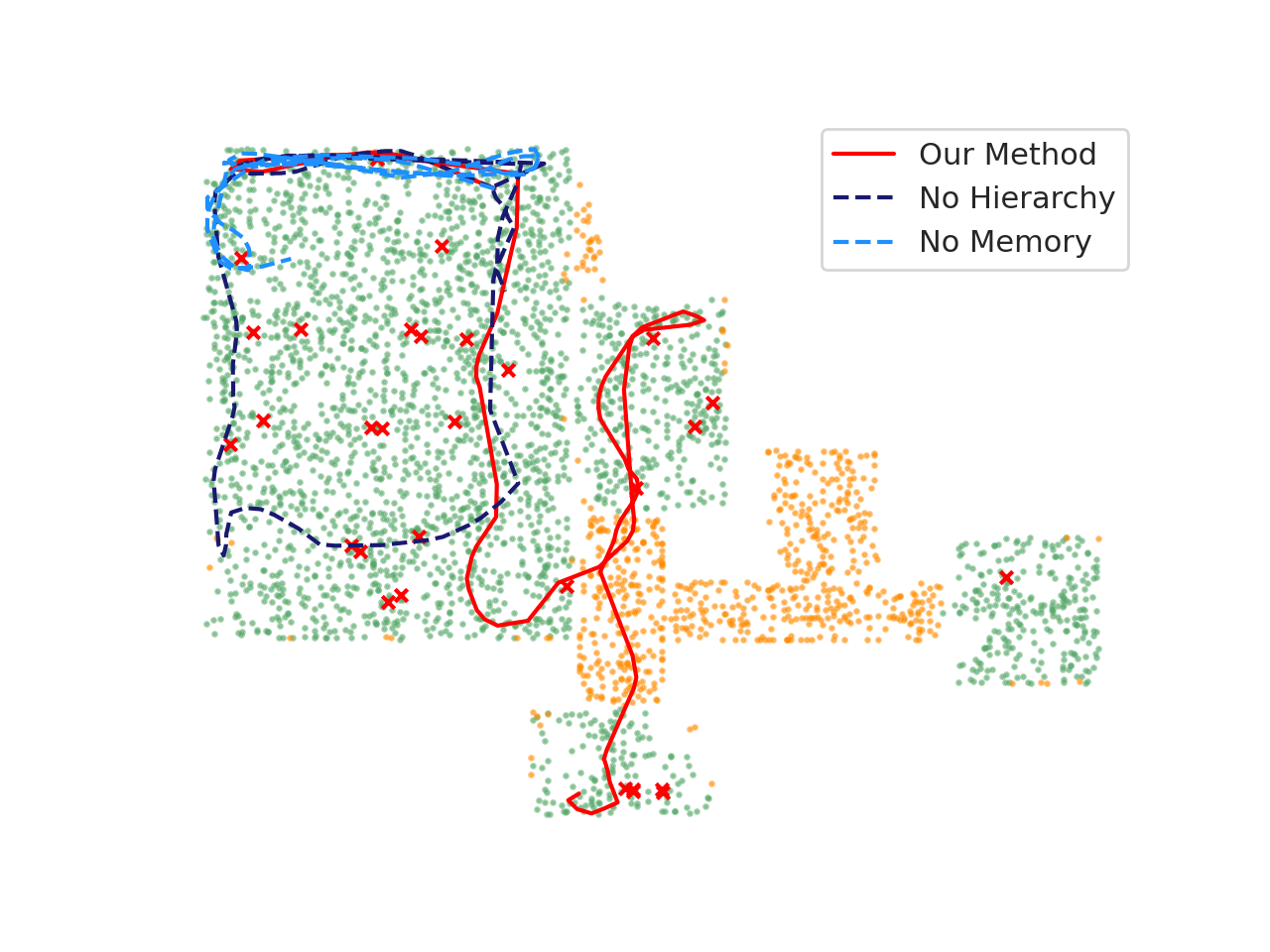}}
	\hfill
	\subfloat[]
	{\includegraphics[width=\figwidth\textwidth, trim=25 10 25 10]{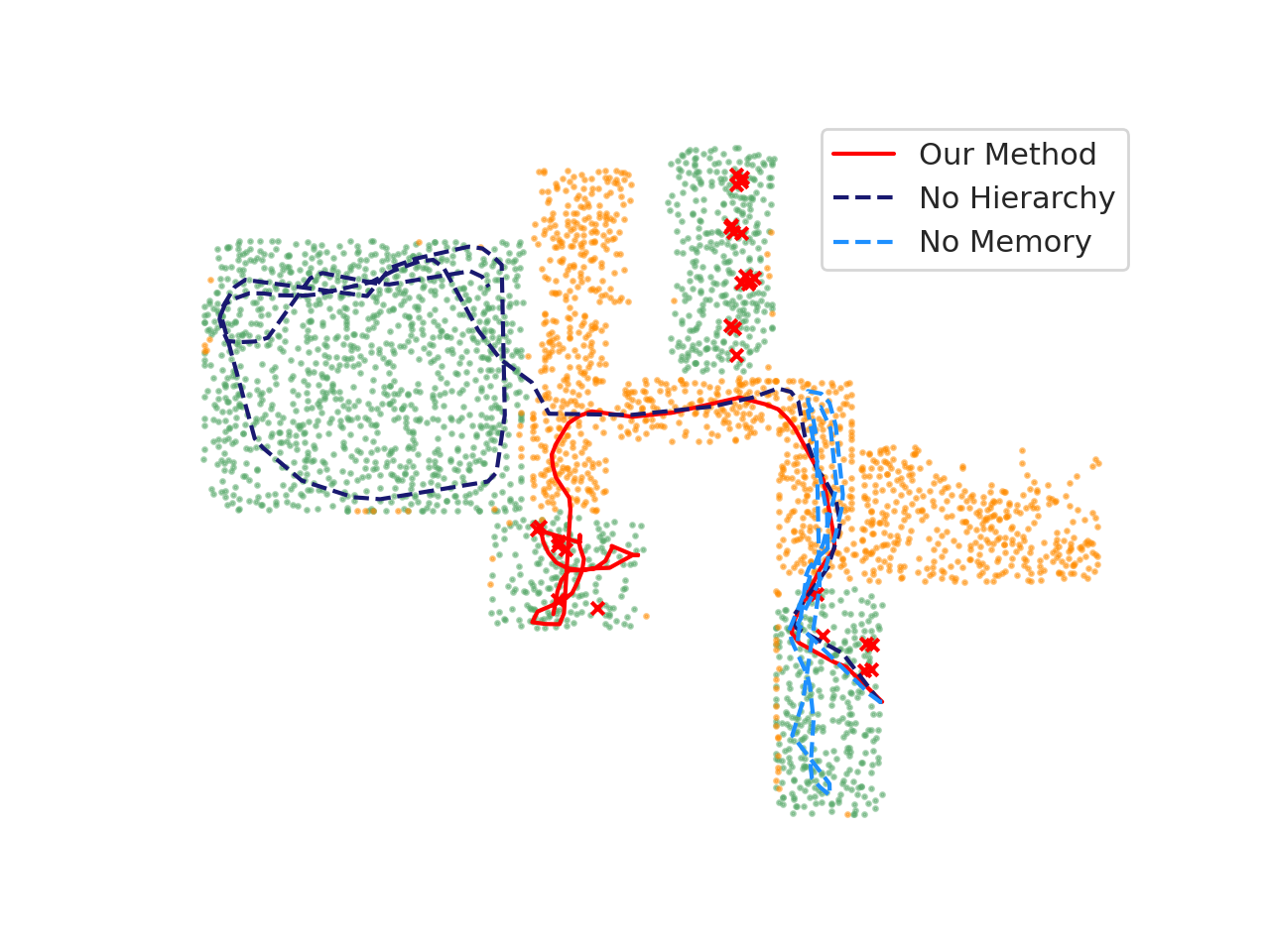}}
	\\
	\subfloat[]
	{\includegraphics[width=\figwidth\textwidth, trim=25 10 25 10]{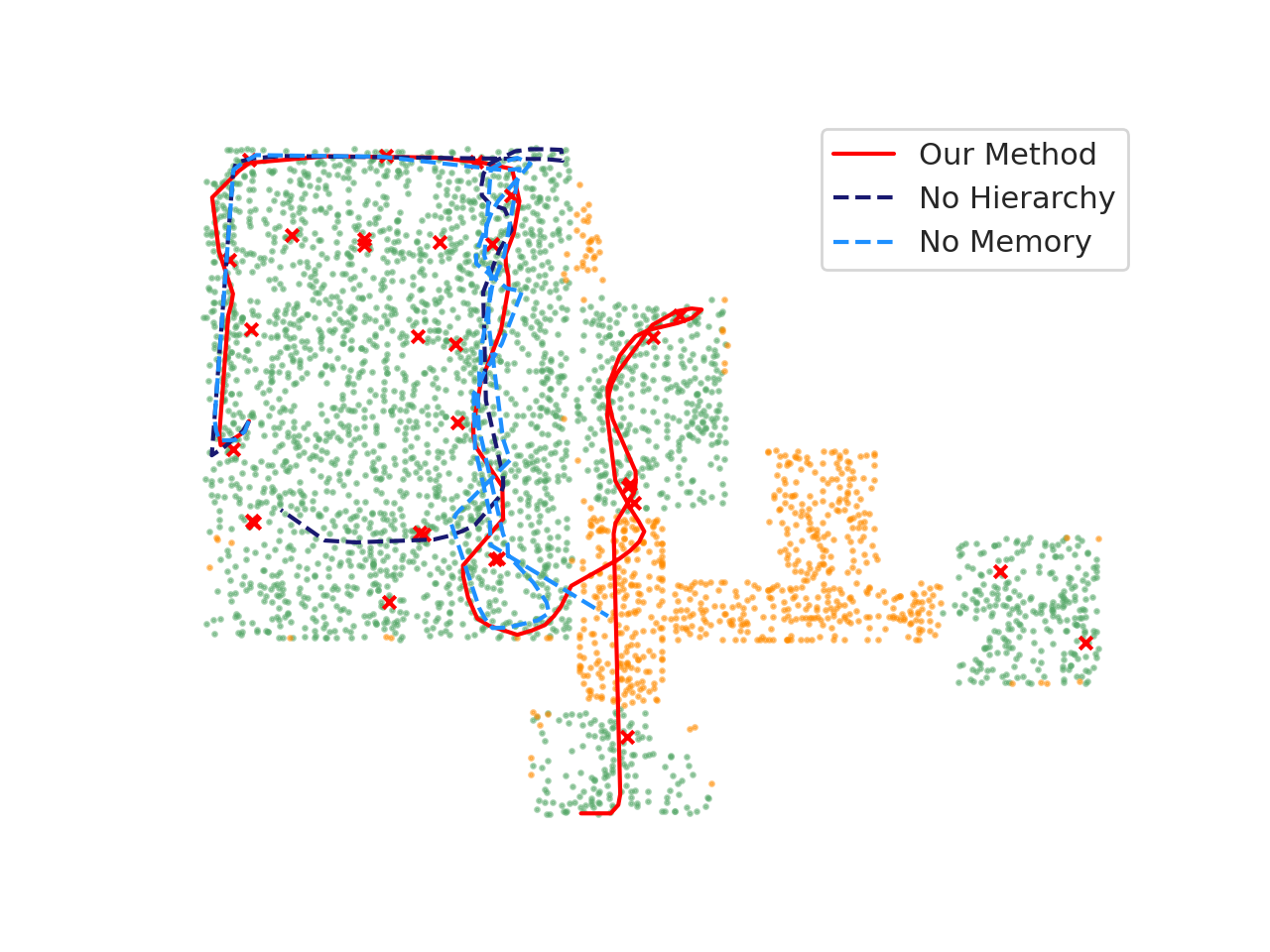}}
	\hfill
	\subfloat[]
	{\includegraphics[width=\figwidth\textwidth, trim=25 10 25 10]{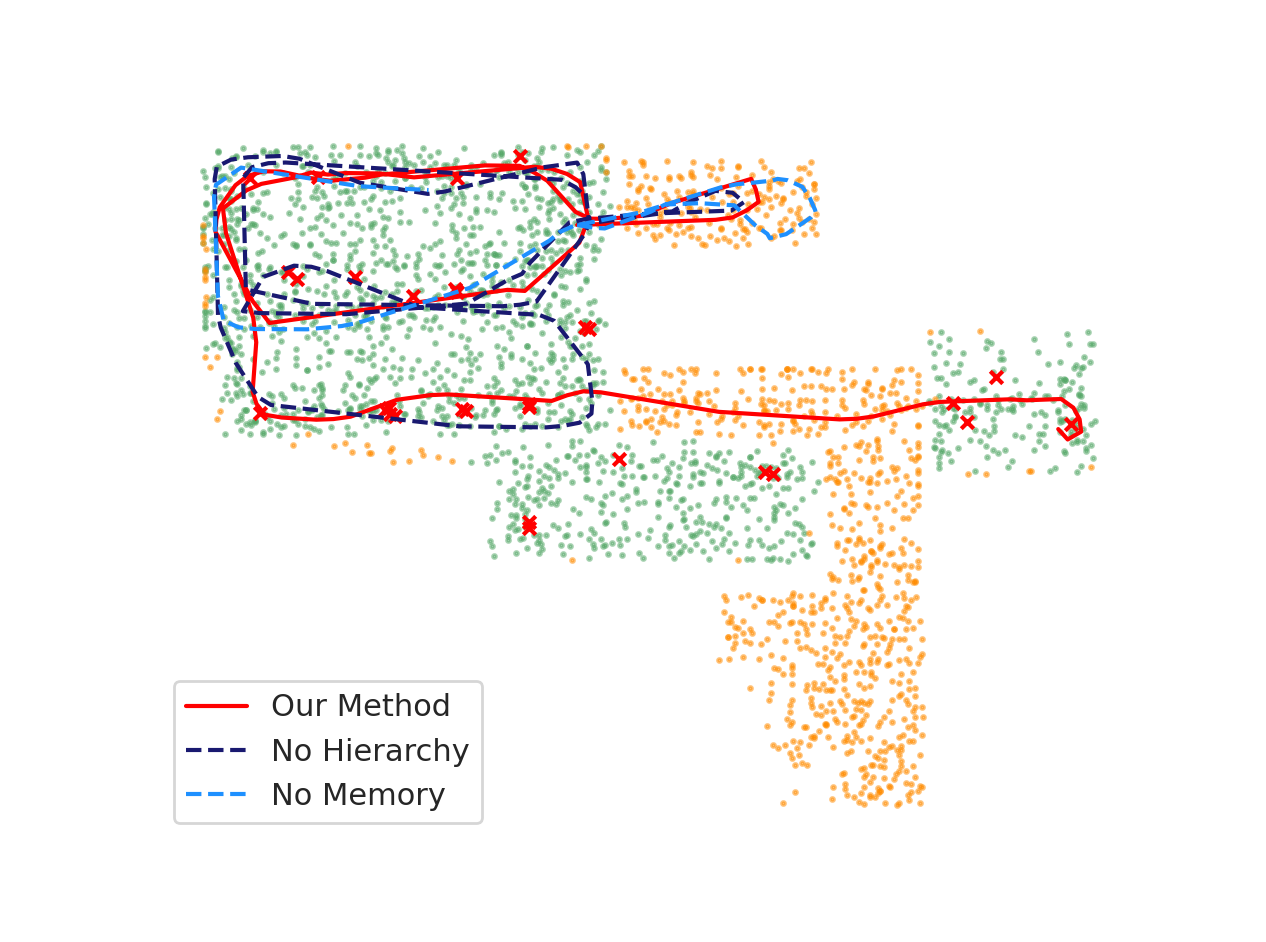}}
	\caption{Example trajectories comparing our method to policies using only the Places layer of the DSG (No Hierarchy) and no explicit memory (No Memory), \highlight{with targets shown in red}.
		Rooms containing targets are labeled green, distractor rooms are orange. Policies without hierarchical information spend less time exploring target rooms.
		Policies without explicit memory spend more time revisiting space. 
		See Section 4.2 of the main paper for quantitative results.}
	\label{fig:dsg-ablation}
\end{figure*}

Figure \ref{fig:baseline-plots} and Figure \ref{fig:dsg-ablation} provide sample trajectories from the multi-object search task described in Section 4.1 of the main paper.
Policies must collect targets placed in semantically distinct rooms (visualized in green).
This task implicitly requires successful policies to avoid visiting distractor rooms (visualized in orange), revisiting space, and colliding with obstacles.

\textbf{Example Trajectories: Comparison with Baselines.}
Figure \ref{fig:baseline-plots} provides visualizations of results from Section 4.2 of the main paper.
Compared to baselines using RGB, RGB-D + Semantics, and ESDF data, our method collects more targets and explores more space.
In Section 4.2 of the main paper we note that ESDF has the most collisions.
Qualitatively, this policy is prone to colliding with object corners (\eg doorways, table corners) and sliding against obstacles (\eg walls).
This collision-prone behavior manifests in shorter trajectories (\eg in Figure~\ref{fig:baseline-plots} (b, e)) as the policy uses episodes steps trying to unsuccessfully maneuver out of a collision, and trajectory segments that closely follow obstacle boundaries (\eg Figure~\ref{fig:baseline-plots} (a)).

\textbf{Example Trajectories: On the Importance of Hierarchy and Explicit Memory.}
Figure \ref{fig:dsg-ablation} provides visualizations of results from Section 4.3 of the main paper.
Our method, which uses observations that provide trajectory history and are derived from multiple layers of the DSG, is compared to policies that use observations formed with only the Places layer (No Hierarchy) and without explicit memory (No Memory). Policies without hierarchical information spend less time exploring target rooms; policies without explicit memory revisit the same areas more often.

\textbf{Training and Inference Times.}
Table \ref{tab:timing} reports training and inference times for methods considered in Section 4.2 (Comparison with Baselines) of the main paper.
Results were generated on a machine with an RTX A6000 GPU and AMD 3990X CPU.
We report average train time per worker step to normalize across varying batch sizes.
Our method takes 5.5 times longer to train than RGB, while this differential drops to 2.8 during inference.

\begin{table*}[!h]
	\centering
	\begin{tabular}{c c c}
		\toprule
		Method            & Mean Train Time (ms per worker step, $\downarrow$) & Mean Inference Time (ms, $\downarrow$) \\
		\midrule
		Our Method        & 30.33                                              & 8.50                                   \\
		RGB-D + Semantics & 7.77                                               & 3.77                                   \\
		RGB               & \textbf{5.56}                                      & \textbf{3.03}                          \\
		ESDF              & 16.38                                              & 4.84                                   \\
		\bottomrule
	\end{tabular}
	\caption{Training and inference times for the results reported in Section 4.2 (Comparison with Baselines) of the main paper.
	}
	\label{tab:timing}
\end{table*}

\section{Effect of Noise on Method Performance}
\label{sec:sim2real}

We study how noise in the DSGs creation impacts the proposed navigation policy.
Since the DSG created by a physical robot is expected to be noisy, this analysis is useful to quantify
a potential \emph{sim-to-real} gap, in preparation for the actual deployment on a real robot.
We develop noise models based on the work by \citet{Rosinol21arxiv-Kimera}, which builds and evaluates DSGs in several indoor spaces. DSGs constructed from real data will not have access to ground-truth localization information, so we perturb node positions. Because the DSG incorporates imperfect semantic segmentation (\eg from a neural network), we inject noise in the DSG's semantic labels. Finally, high-level information (\ie Room nodes) will not be immediately available, but must be inferred only after a sufficient amount of low-level information (\ie Places nodes) has been observed. We model this estimation process by adding latency to Room node observations.
These experiments use the policy trained with noiseless observation as described Section 4.1 of the main paper, and inject noise during evaluation.

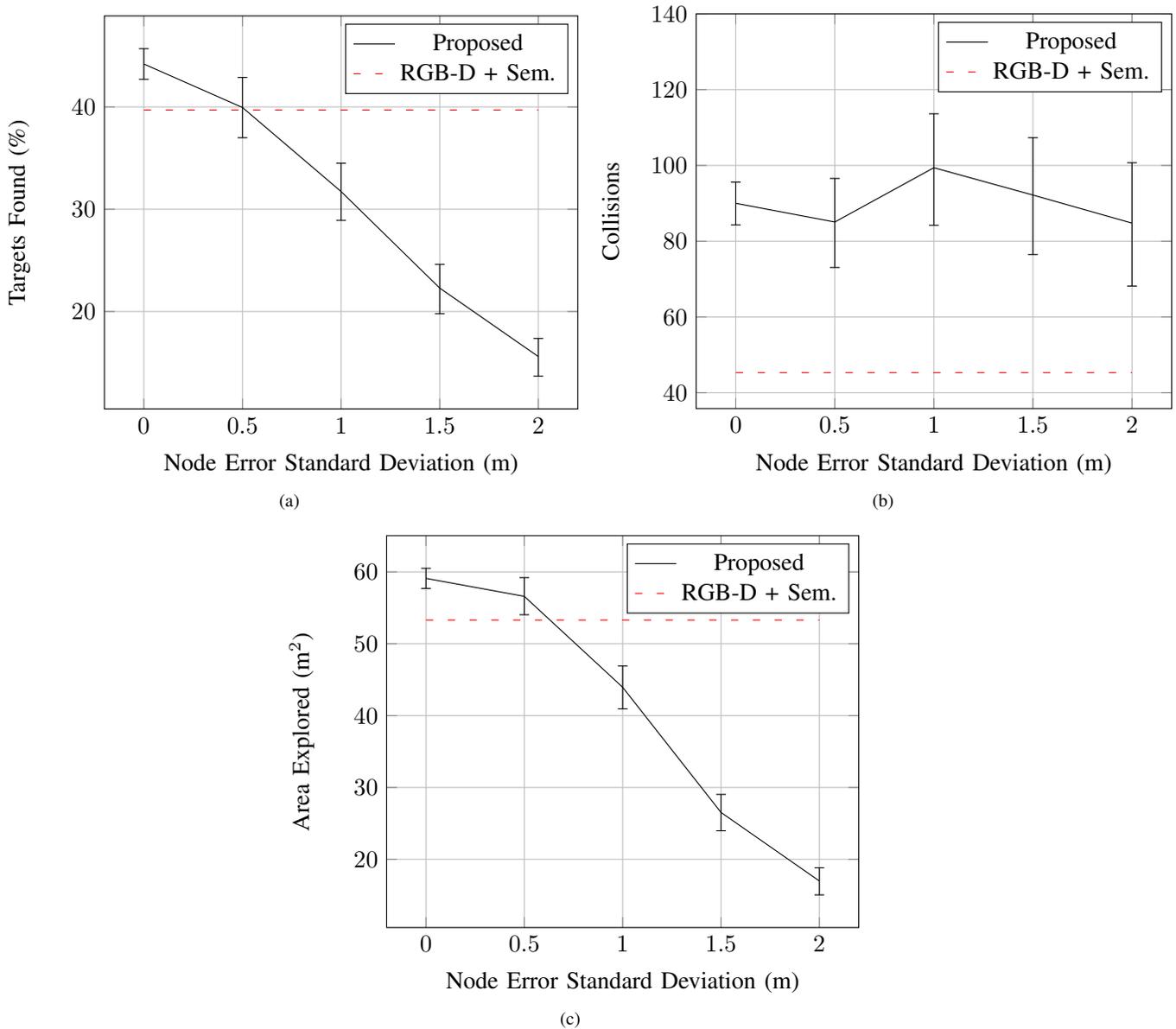
\begin{figure*}[!h]
    \centering
    \subfloat[]
    {
        \begin{adjustbox}{width=0.5\textwidth}
            \begin{tikzpicture}
                \begin{axis}[
                        xlabel=Node Error Standard Deviation (m),
                        ylabel=Targets Found (\%),
                        grid=major,
                    ]
                    \addplot [error bars/.cd,y dir=both, y explicit] coordinates {
                            (0, 44.2)  -= (0, 1.5)  += (0, 1.5)
                            (0.5, 39.95) -= (0, 2.96) += (0, 2.937)
                            (1, 31.73) -= (0, 2.82) += (0, 2.775)
                            (1.5, 22.29) -= (0, 2.51) += (0, 2.32)
                            (2, 15.59) -= (0, 1.91) += (0, 1.777)
                        };
                    \addlegendentry{Proposed}

                    \addplot [no markers, loosely dashed, red] coordinates {(0, 39.7) (2, 39.7)};
                    \addlegendentry{RGB-D + Sem.}
                \end{axis}
            \end{tikzpicture}
        \end{adjustbox}
    }
    \subfloat[]
    {
        \begin{adjustbox}{width=0.5\textwidth}
            \begin{tikzpicture}
                \begin{axis}[
                        xlabel=Node Error Standard Deviation (m),
                        ylabel=Collisions,
                        grid=major,
                        ymax=140,
                    ]
                    \addplot [error bars/.cd,y dir=both, y explicit] coordinates {
                            (0, 90.0) -= (0, 5.7)  += (0, 5.6)
                            (0.5, 85.06) -= (0, 12.0) += (0, 11.51)
                            (1, 99.39) -= (0, 15.2) += (0, 14.24)
                            (1.5, 92.18) -= (0, 15.7) += (0, 15.13)
                            (2, 84.77) -= (0, 16.6) += (0, 15.93)
                        };
                    \addlegendentry{Proposed}

                    \addplot [no markers, loosely dashed, red] coordinates {(0, 45.3) (2, 45.3)};
                    \addlegendentry{RGB-D + Sem.}
                \end{axis}
            \end{tikzpicture}
        \end{adjustbox}
    }
    \\
    \subfloat[]
    {
        \begin{adjustbox}{width=0.5\textwidth}
            \begin{tikzpicture}
                \begin{axis}[
                        xlabel=Node Error Standard Deviation (m),
                        ylabel=Area Explored (m$^2$),
                        grid=major,
                    ]
                    \addplot [error bars/.cd,y dir=both, y explicit] coordinates {
                            (0, 59.1)  -= (0, 1.4) += (0, 1.4)
                            (0.5, 56.6) -= (0, 2.56) += (0, 2.603)
                            (1, 43.94) -= (0, 3.0) += (0, 2.965)
                            (1.5, 26.52) -= (0, 2.55) += (0, 2.515)
                            (2, 16.98) -= (0, 1.93) += (0, 1.837)
                        };
                    \addlegendentry{Proposed}

                    \addplot [no markers, loosely dashed, red] coordinates {(0, 53.3) (2, 53.3)};
                    \addlegendentry{RGB-D + Sem.}

                \end{axis}
            \end{tikzpicture}
        \end{adjustbox}
    }
    \caption{\highlight{Effect on node position noise. Noise is sampled from a Gaussian Distribution with zero-mean and varying standard deviation. Increased position noise leads to decrease in both targets found and area explored.}\label{fig:pos_noise}}
\end{figure*}

\textbf{Effect of Position Error.}
We first perturb node positions with Gaussian noise. We use zero-mean noise with standard deviations (SD) sampled at 0.5\,m intervals between 0\,m and 2\,m.  Per-node perturbation is generated at the start of the episode and held constant throughout. Figure~\ref{fig:pos_noise} shows the effect of noise on the number of targets found, collisions, and area explored. The policy collects 40\% of the targets with SD = 0.5\,m, which is slightly higher than the RGB-D + Semantics model using ground truth depth and pose.
We note that this level of noise is significantly higher than the expected position errors reported in~\cite{Rosinol21arxiv-Kimera}.
Interestingly, the decrease in found targets is correlated with area explored rather than collisions.
Qualitative analysis shows that with increasing levels of noise the policy acts as if it were near obstacles (\eg performing tight maneuvers), even in open space. The resulting trajectories are inefficient, thus explore less area. Such behavior suggests that the policy uses immediate node positions to infer free-space; when these positions are perturbed, the policy assumes there are nearby obstacles that it must avoid.

\begin{figure*}[!h]
    \centering
    \subfloat[]
    {
        \begin{adjustbox}{width=0.5\textwidth}
            \begin{tikzpicture}
                \begin{axis}[
                        xlabel=Percentage of Corrupted Labels (\%),
                        ylabel=Targets Found (\%),
                        grid=major,
                    ]
                    \addplot [error bars/.cd,y dir=both, y explicit] coordinates {
                            (0, 44.67)-= (0, 2.9)  += (0, 2.0)
                            (20, 43.01)-= (0, 2.83)  += (0, 2.76)
                            (40, 41.96)-= (0, 2.95)  += (0, 2.867)
                            (60, 40.18)-= (0, 3.05)  += (0, 2.978)
                            (80, 39.31)-= (0, 2.76)  += (0, 2.787)
                        };
                    \addlegendentry{Proposed}

                    \addplot [no markers, loosely dashed, red] coordinates {(0, 39.7) (80, 39.7)};
                    \addlegendentry{RGB-D + Sem.}

                    \addplot [no markers, densely dashed, blue] coordinates {(0, 41.9) (80, 41.9)};
                    \addlegendentry{No Hierarchy}

                \end{axis}
            \end{tikzpicture}
        \end{adjustbox}
    }
    \subfloat[]
    {
        \begin{adjustbox}{width=0.5\textwidth}
            \begin{tikzpicture}
                \begin{axis}[
                        xlabel=Percentage of Corrupted Labels (\%),
                        ylabel=Collisions,
                        grid=major,
                        ymax=140,
                    ]
                    \addplot [error bars/.cd,y dir=both, y explicit] coordinates {
                            (0, 90) -= (0, 5.7) += (0, 5.6)
                            (20, 80.61) -= (0, 9.85) += (0, 9.608)
                            (40, 89.5) -= (0, 11.2) += (0, 10.41)
                            (60, 95.07) -= (0, 12.6) += (0, 12.14)
                            (80, 90.12) -= (0, 11.3) += (0, 10.76)
                        };

                    \addlegendentry{Proposed}

                    \addplot [no markers, loosely dashed, red] coordinates {(0, 45.3) (80, 45.3)};
                    \addlegendentry{RGB-D + Sem.}

                    \addplot [no markers, densely dashed, blue] coordinates {(0, 91.9) (80, 91.8)};
                    \addlegendentry{No Hierarchy}
                \end{axis}
            \end{tikzpicture}
        \end{adjustbox}
    }
    \\
    \subfloat[]
    {
        \begin{adjustbox}{width=0.5\textwidth}
            \begin{tikzpicture}
                \begin{axis}[
                        xlabel=Percentage of Corrupted Labels (\%),
                        ylabel=Area Explored (m$^2$),
                        grid=major,
                        ymax=64,
                    ]
                    \addplot[error bars/.cd,y dir=both, y explicit] coordinates {
                            (0, 59.1) -= (0, 1.4) += (0, 1.4)
                            (20, 59.08) -= (0, 2.36) += (0, 2.508)
                            (40, 57.69) -= (0, 2.54) += (0, 2.58)
                            (60, 55.79) -= (0, 2.9) += (0, 3.089)
                            (80, 57.06) -= (0, 2.66) += (0, 2.648)
                        };
                    \addlegendentry{Proposed}

                    \addplot [no markers, loosely dashed, red] coordinates {(0, 53.3) (80, 53.3)};
                    \addlegendentry{RGB-D + Sem.}

                    \addplot [no markers, densely dashed, blue] coordinates {(0, 58.3) (80, 58.3)};
                    \addlegendentry{No Hierarchy}
                \end{axis}
            \end{tikzpicture}
        \end{adjustbox}
    }
    \caption{\highlight{Effect of semantic label noise on method performance. The number of targets found steadily decreases as semantic noise increases, while collisions and area explored remain relatively stable.}\label{fig:semantic_noise}}
\end{figure*}
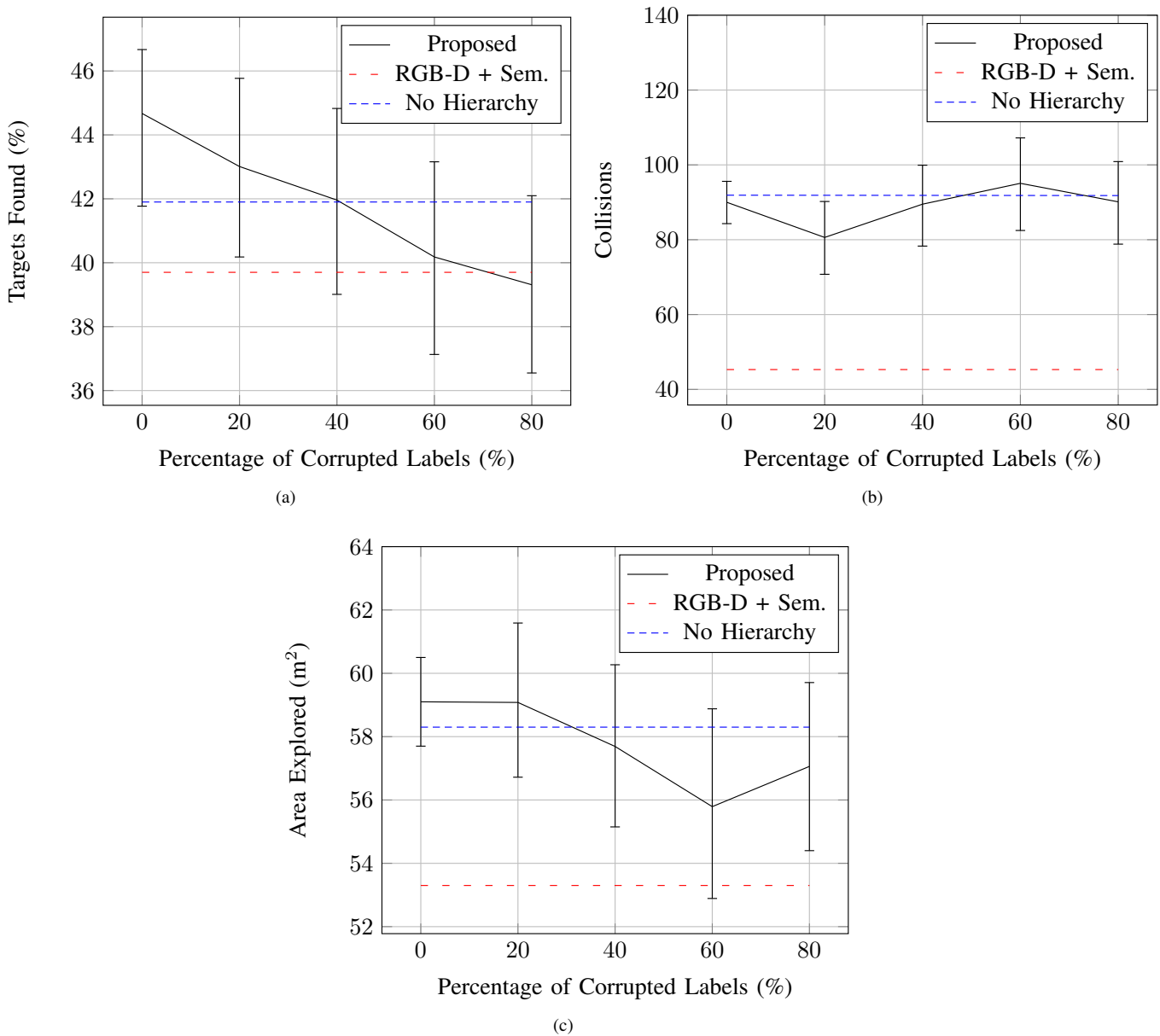

\textbf{Effect of Semantic Noise.}
We next introduce semantic noise by randomly changing a percentage of DSG labels; we start at 0\% then increment by 20\%. Similarly to the above experiment, noise is generated at the start of an episode then held constant throughout. Results are shown in Figure~\ref{fig:semantic_noise}. The number of targets found steadily decreases with increasing semantic noise, while collisions and area explored remain relatively steady. After roughly 40\% semantic perturbation, the number of targets found drops below No Hierarchy performance, which doesn't rely on explicit semantic information. This trend suggests that after a certain level of noise, imperfect semantic labels mislead the policy rather than help guide its navigation objectives; it also
demonstrates that our policy indeed learns how to leverage semantic information.

\newcommand{\semanticnoiselabel}{Semantic Noise (\%)}

\textbf{Effect of Room Node Delay.}
As noted above, real-time DSG construction must infer high-level information from low-level observations, and in practice this inference relies on a sufficient amount of low-level information being available; for instance, a robot can infer the semantic label of a room only after seeing some low-level aspect of the room (\eg places, objects) .
To study the effect of potential delays in the inference of high-level layers in the scene graph, we run an experiment where Room nodes are added to the DSG only after 50\% of its child Place nodes have been identified. Table~\ref{tab:delayed_rooms} compares delayed room observations with the original method and No Hierarchy results. While delaying room observations does impact performance, such information still provides benefit over a flat representation.

\begin{table*}[!h]
    \centering
    \begin{tabular}{llll}
        \toprule
        Method        & Targets Found (\%, $\uparrow$) & Collisions ($\downarrow$)  & Area Explored (m$^2$, $\uparrow$) \\
        \midrule
        Proposed      & \textbf{44.2} (42.7, 45.7)     & \textbf{90.0} (84.3, 95.6) & \textbf{59.1} (57.7, 60.5)        \\
        Delayed Rooms & 43.0 (40.2, 45.9)              & 98.0 (85.5, 109.7)         & 57.7 (55.0, 60.6)                 \\
        No Hierarchy  & 41.9 (40.5, 43.3)              & 91.9 (85.6, 98.3)          & 58.3 (56.8, 59.7)                 \\
        \bottomrule
    \end{tabular}
    \caption{\highlight{Effect of Room node observation delay. ``Delayed Rooms'': rooms are only added to the DSG after 50\% of the child Places nodes have been observed (the remaining rows are the same as Table~2 in the main paper). Delaying Room nodes leads to a drop in targets found, but still provides an advantage over not using hierarchical information.}\label{tab:delayed_rooms}}
\end{table*}

\end{document}